\documentclass[pdflatex,sn-basic,iicol,oneside,openany]{sn-jnl}% Basic Springer Nature Reference Style/Chemistry Reference Style
\usepackage{graphicx}%
\usepackage{multirow}%
\usepackage{amsmath,amssymb,amsfonts}%
\usepackage{amsthm}%
\usepackage{mathrsfs}%
\usepackage[title]{appendix}%
\usepackage{xcolor}%
\usepackage{textcomp}%
\usepackage{natbib} 
\usepackage{booktabs}%
\usepackage{listings}%

\usepackage{balance}
\usepackage{array}
\usepackage{stfloats}
\usepackage{url}
\usepackage{verbatim}
\usepackage{multicol}
\usepackage{subfloat}
\usepackage{colortbl}
\usepackage[ruled]{algorithm2e}

\usepackage{bbding}
\usepackage{hyperref}
\usepackage{ulem}
\usepackage{makecell}
\usepackage{adjustbox}

\theoremstyle{thmstyleone}%
\theoremstyle{thmstyletwo}%

\theoremstyle{thmstylethree}%

\begin{document}

\title[Reflections Unlock: Geometry-Aware Reflection Disentanglement in 3D Gaussian Splatting for Photorealistic Scenes Rendering]{Reflections Unlock: Geometry-Aware Reflection Disentanglement in 3D Gaussian Splatting for Photorealistic Scenes Rendering}

\author[1]{\fnm{Jiayi} \sur{Song}}\email{22307130359@m.fudan.edu.cn}
\equalcont{These authors contributed equally to this work.}

\author[1]{\fnm{Zihan} \sur{Ye}}\email{22300240027@m.fudan.edu.cn}
\equalcont{These authors contributed equally to this work.}

\author[1]{\fnm{Qingyuan} \sur{Zhou}}\email{zhouqy23@m.fudan.edu.cn}
\equalcont{These authors contributed equally to this work.}

\author[1]{\fnm{Weidong} \sur{Yang}}\email{wdyang@fudan.edu.cn}

\author*[2]{\fnm{Ben} \sur{Fei}}\email{benfei@cuhk.edu.hk}

\author[1]{\fnm{Jingyi} \sur{Xu}}\email{jy22@m.fudan.edu.cn}

\author[3]{\fnm{Ying} \sur{He}}\email{yhe@ntu.edu.sg}

\author[2]{\fnm{Wanli} \sur{Ouyang}}\email{wlouyang@ie.cuhk.edu.hk}

\affil[1]{\orgdiv{School of Computer Science}, \orgname{Fudan University}, \orgaddress{\city{Shanghai}, \country{China}}}

% \affil*[2]{\orgname{Shanghai Artificial Intelligence Laboratory}, \orgaddress{\city{Shanghai}, \country{China}}}

\affil[2]{\orgdiv{Department of Information Engineering}, \orgname{The Chinese University of Hong Kong}, \orgaddress{\city{Hong Kong}, \country{China}}}

\affil[3]{\orgdiv{College of Computing and Data Science}, \orgname{Nanyang Technological University}, \orgaddress{\country{Singapore}}}

% \affil[5]{\orgdiv{School of Electronics Information and Electrical Engineering}, \orgname{Shanghai Jiao Tong University}, \orgaddress{\city{Shanghai},  \country{China}}}

% \vspace{-10em}

\abstract{
   Accurately rendering scenes with reflective surfaces remains a significant challenge in novel view synthesis, as existing methods like Neural Radiance Fields (NeRF) and 3D Gaussian Splatting (3DGS) often misinterpret reflections as physical geometry, resulting in degraded reconstructions. Previous methods rely on incomplete and non-generalizable geometric constraints, leading to misalignment between the positions of Gaussian splats and the actual scene geometry. When dealing with real-world scenes containing complex geometry, the accumulation of Gaussians further exacerbates surface artifacts and results in blurred reconstructions. To address these limitations, in this work, we propose Ref-Unlock, a novel geometry-aware reflection modeling framework based on 3D Gaussian Splatting, which explicitly disentangles transmitted and reflected components to better capture complex reflections and enhance geometric consistency in real-world scenes. Our approach employs a dual-branch representation with high-order spherical harmonics to capture high-frequency reflective details, alongside a reflection removal module providing pseudo reflection-free supervision to guide clean decomposition. Additionally, we incorporate pseudo-depth maps and a geometry-aware bilateral smoothness constraint to enhance 3D geometric consistency and stability in decomposition. Extensive experiments demonstrate that Ref-Unlock significantly outperforms classical GS-based reflection methods and achieves competitive results with NeRF-based models, while enabling flexible vision foundation models (VFMs) driven reflection editing. Our method thus offers an efficient and generalizable solution for realistic rendering of reflective scenes. Our code is available at \href{https://ref-unlock.github.io/}{https://ref-unlock.github.io/}.
   }

\keywords{Novel view synthesis, real-time rendering, reflection modeling, 3D Gaussian Splatting.\vspace{10em}}

\maketitle

\let\cleardoublepage\clearpage
\section{Introduction}
\label{sec:Introduction}
Novel view synthesis from multi-view images has emerged as a pivotal research direction in computer vision and graphics, driven by its wide applications in virtual reality~\citep{wang2024slide}, augmented reality~\citep{zuo2025fmgs}, and 3D reconstruction~\citep{liu2025image}.
% Neural Radiance Fields (NeRF)~\citep{reiser2021kilonerf} pioneered this field by introducing an implicit scene representation capable of photo-realistic, view-dependent rendering.
The advent of Neural Radiance Fields (NeRF)\citep{reiser2021kilonerf} has marked a significant breakthrough in scene representation by enabling photorealistic image synthesis with view-dependent effects.
% It employs a fully connected neural network to map spatial coordinates and viewing directions to color and density, which are integrated via volumetric rendering to produce final pixel color~\citep{jain2024learnin}.
This approach models scenes implicitly using a multi-layer perceptron that translates 3D positions and camera directions into color and density, which are then composited using volumetric rendering techniques\citep{jain2024learning}.
% Despite its remarkable performance, NeRF and its variants~\citep{deng2022depth,barron2021mip,pumarola2021d} are constrained by computational complexity and the time-consuming optimization process.
Although NeRF-based methods~\citep{deng2022depth,barron2021mip,pumarola2021d} achieves impressive visual fidelity, its practical use is limited due to high computational demands and the extensive time required for model optimization.
In contrast, 3D Gaussian Splatting (3DGS)~\citep{kerbl20233d} offers an explicit scene representation using anisotropic 3D Gaussians.
It initializes Gaussians from a Structure-from-Motion (SfM)~\citep{schonberger2016structure} point cloud and projects them onto the 2D image plane during rendering, with colors computed from Spherical Harmonics (SH) parameters and images synthesized via differentiable rasterization.
% With efficient optimization and rendering, 3DGS outperforms NeRF in real-time applications while maintaining a comparative rendering quality.
By leveraging fast training and rendering strategies, 3DGS achieves real-time performance and delivers rendering results comparable to those of NeRF.
% On the contrary, the newly proposed 3D Gaussian Splatting (3DGS)~\citep{kerbl20233d} has emerged as a promising alternative approach, which represents scenes explicitly with anisotropic 3D Gaussians.

%each Gaussian ball is defined by the mean position, covariance matrix for the shape control, opacity and  spherical harmonics parameters for modeling view-dependent color.
% 3DGS first initializes Gaussians parameters from the Structure-from-motion (SfM)~\citep{schonberger2016structure} point cloud. During the rendering process, these Gaussians are splatted to the 2D image plane, and their colors are computed from the Spherical Harmonics (SH) parameters. Then, point-based differentiable rasterization is used to render the image from novel viewpoints. 

% Teaser Image
% \vspace{-0.5cm}
\begin{figure}[htbp]
    \centering
    \includegraphics[width=\linewidth]{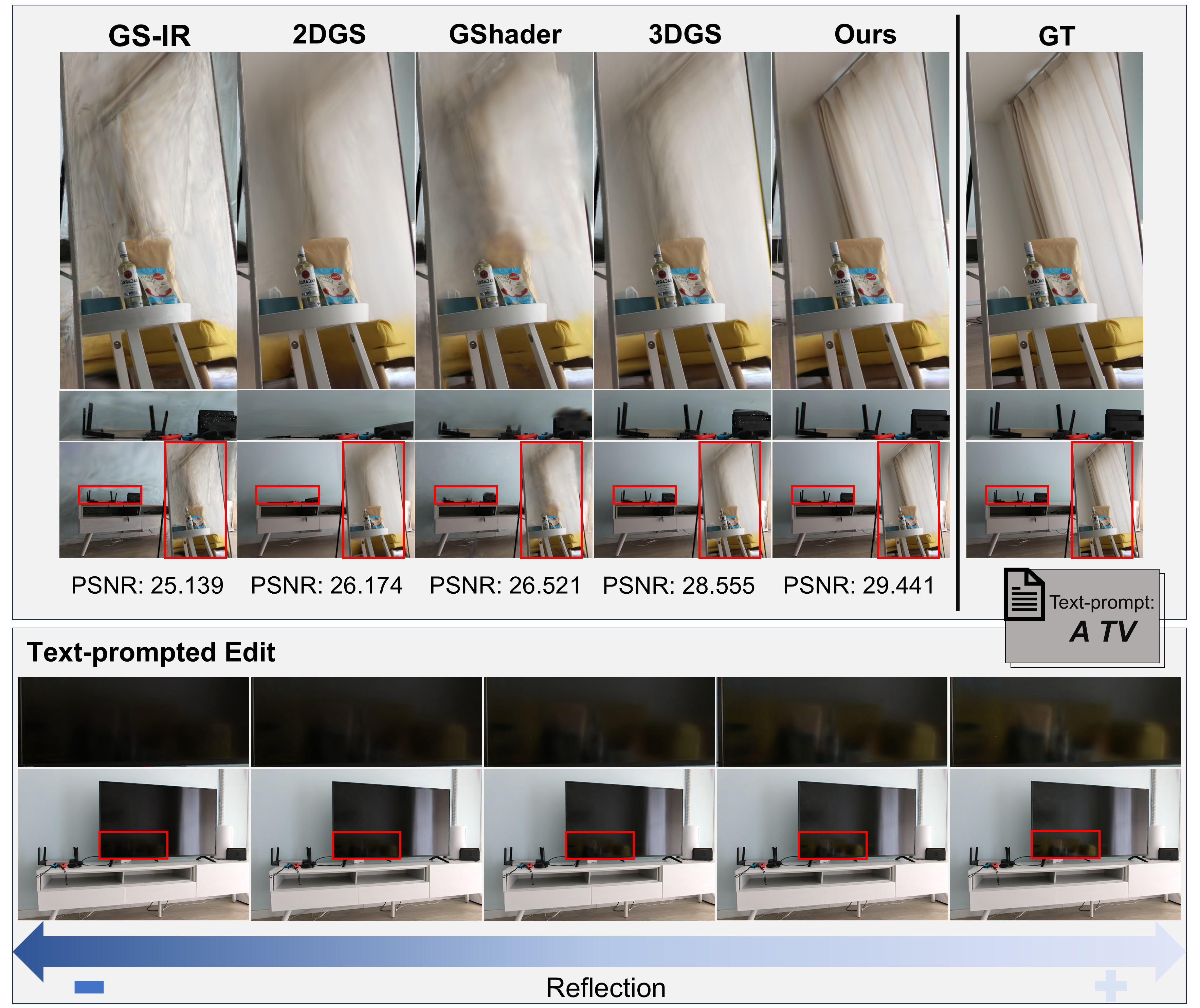}
    % \vspace{-0.5cm}
    \caption{\textbf{Novel view synthesis with Ref-Unlock, fulfilling better reflections modeling than other methods.} The top row compares rendering quality of various methods (GS-IR~\citep{liang2024gs}, 2DGS~\citep{huang20242d}, GShader~\citep{jiang2024gaussianshader}, 3DGS~\citep{kerbl20233d}, Ours) against the ground truth (GT), with PSNR scores provided for quantitative evaluation. 
    The bottom row shows the VFMs-driven reflection editing with varying reflection strength scales using our method.}
    \label{fig:teaser}
\end{figure}

% Nonetheless, both NeRF and 3DGS fall short of accurately reconstructing scenes that feature reflective surfaces such as mirrors, glass, and screens~\citep{fei20243d,zhou2025mgsr}.
However, NeRF and 3DGS both struggle to faithfully reconstruct environments containing reflective materials—such as mirrors, transparent glass, or display screens~\citep{fei20243d,zhou2025mgsr}.
% This limitation arises from the absence of explicit modeling for strong reflection components, leading to failures in handling surfaces like mirrors, glass, and screens.
This challenge stems from their lack of explicit mechanisms to model strong specular reflections, which hampers their ability to handle such surfaces effectively.
% As a consequence, these methods often misinterpret reflected content as physically existing objects, severely compromising the fidelity and consistency of the rendered scenes.
As a result, such approaches frequently mistake reflections for actual scene geometry, leading to significant degradation in rendering accuracy and visual coherence.
Furthermore, since reflections change with the viewpoint, this misrepresentation leads to inconsistencies across multiple views~\citep{yao2024reflective}. 
These inconsistencies, in turn, introduce conflicts during optimization, ultimately resulting in blurred and degraded reconstructions~\citep{verbin2022ref}.

% To address this challenge, several methods have been proposed for improving scene reconstruction in the presence of reflections.
% \textcolor{blue}{NeRF-based}
To address this issue, NeRF-based methods like Mirror-NeRF~\citep{zeng2023mirror} and TraM-NeRF~\citep{vanholland2023tramnerf} are specifically devised for mirror reflections. 
% cast their eyes on handling mirror reflections. 
% Mirror-NeRF~\citep{zeng2023mirror} parameterizes spatial points with reflection probability and uses Whitted Ray Tracing~\citep{whitted2005improved} to blend pixel colors from direct and reflected rays based on this probability.
% TraM-NeRF~\citep{vanholland2023tramnerf} models mirror reflections with physically plausible materials and integrates volume rendering with Monte Carlo methods to estimate reflected radiance.
In contrast, methods like Ref-NeRF~\citep{verbin2022ref} and NeRFReN~\citep{guo2022nerfren} extend the NeRF framework to handle more diverse types of reflections, including those on non-ideal surfaces.
% Ref-NeRF ~\citep{verbin2022ref} introduces a reflection-aware volume rendering technique, explicitly modeling reflections on glossy surfaces. This allows the network to integrate both direct and reflected lighting components, offering a more realistic representation of reflections in complex scenes.
% NeRFReN~\citep{guo2022nerfren} partitions the scene into multiple subspaces, enabling separate handling of different reflection types (e.g., diffuse and specular). This scene partitioning strategy makes the model more efficient in rendering complex reflections and increases its ability to handle diverse reflective phenomena.
% Despite these advancements, these methods inevitably inherit the limitation of slow optimization as an unresolved challenge~\citep{lin2024fastsr}.
Although notable progress has been made, these approaches continue to suffer from the persistent issue of time-consuming optimization, which remains an open problem~\citep{lin2024fastsr}.
% \textcolor{blue}{GS-based towards mirror with masks}
% The Mirror3DGS~\citep{meng2024mirror} and MirrorGaussian~\citep{liu2024mirrorgaussian} methods, which are based on 3DGS, utilize the plane equation to represent the mirror and exploit the symmetry to combine the real-world image with its mirrored counterpart for the ultimate outcome. 
% Nonetheless, these approaches necessitate accurate mirror masks to assist in the mirror parameterization and solely focus on modeling specular reflections (Table.~\ref{tab:method_com}).
% \textcolor{blue}{GS-based BRDF. BRDF vs only SH.}
To overcome this, GS-based methods have emerged as a promising alternative. 
Among these, Physically Based Rendering (PBR) methods, including Bidirectional Reflectance Distribution Function (BRDF)-based approaches, focus on modeling reflections based on physical principles, such as how light interacts with surface geometry and material properties. 
Some approaches enhance appearance modeling through PBR-based shading and color decomposition~\citep{jiang2024gaussianshader}, while others integrate normals, materials, and lighting to enable realistic relighting~\citep{gao2023relightable, liang2024gs}. Additionally, advanced shading strategies such as deferred rendering and reflection-guided normal optimization have been proposed to improve specular realism~\citep{ye20243d}.
% GShader~\citep{jiang2024gaussianshader} apply simplified BRDF shading functions on 3D Gaussians to enhance neural rendering in scenes with reflective surfaces, focusing on decomposing color attributes into components such as diffuse, specular, and residual color for modeling reflective surfaces.
% R3DG~\citep{gao2023relightable} extends 3D Gaussian Splatting with embedded normals, BRDF parameters, and incident lighting, and uses point-based ray tracing with BVH for efficient visibility handling, enabling realistic relighting.
% GS-IR~\citep{liang2024gs} extends 3D Gaussian Splatting with staged optimization to jointly recover geometry, material, and lighting under unknown illumination, incorporating normal estimation and occlusion modeling within a full PBR pipeline.
% 3DGS-DR~\citep{ye20243d} leverages deferred shading with per-pixel reflection gradients to optimize surface normals during 3D Gaussian Splatting for improved specular reflection rendering.

While recent GS-based PBR methods achieve impressive results, their geometric constraints—particularly regarding depth estimation—lack applicability across various scenes.
This often leads to misalignment between the positions of Gaussian primitives and the underlying scene geometry~\citep{huang20242d}. 
In scenes with complex geometry, such as locally intricate surfaces—such as locally intricate surfaces like the curtains reflected in the mirror (Fig.~\ref{fig:teaser}), the accumulation of misaligned Gaussians further amplifies surface artifacts and results in noticeable blurring.
% While these GS-based PBR methods achieve impressive results, they are generally dependent on the accurate modeling of scene-specific properties, such as lighting distribution and surface normals.
% As a result, these methods can be sensitive to the particular characteristics of the scene being reconstructed. 
% While these methods perform well on certain datasets, their effectiveness tends to diminish on more general scenes. 
As shown in Fig.~\ref{fig:teaser}, existing methods such as GS-IR~\citep{liang2024gs} and GShader~\citep{jiang2024gaussianshader} yield suboptimal performance compared to 3DGS~\citep{kerbl20233d} on the RFFR dataset~\citep{guo2022nerfren}. 
Similarly, models like R3DG~\citep{gao2023relightable} and 3DGS-DR~\citep{ye20243d} also encounter difficulties in rendering these scenes consistently. 
These methods typically perform well on curated datasets but exhibit limited applicability to more diverse and real-world environments that contain complex and fine-grained geometric structures.
% These methods often demonstrate strong performance on carefully designed or curated datasets, but tend to generalize less effectively to more diverse and complex real-world scenes.
% To improve their generalization, some models~\citep{ye20243d,yao2024reflective} often require careful parameter tuning for different instances even within the same dataset,
To enhance generalization, some approaches~\citep{ye20243d,yao2024reflective} require case-specific parameter tuning, while others~\citep{liang2024gs,gao2023relightable} adopt staged training strategies—typically stabilizing geometry first before refining material and lighting properties.
% Others~\citep{liang2024gs,gao2023relightable} rely on staged training strategies, first stabilizing geometry before refining material and lighting.

% In contrast, SH-only methods like 3DGS~\citep{kerbl20233d}, due to their ability to efficiently model spherical variations in color and lighting, are less reliant on scene-specific properties like surface normals or lighting distributions. This makes them inherently more robust, as they do not require extensive parameter adjustments or staged training, offering a more stable and consistent approach across different scenes. Also, SH naturally fits the ``massive small particles and rapid blending'' strategy of Gaussian Splatting, emphasizing local approximations while maintaining computational efficiency.

In contrast, SH-only methods like 3DGS~\citep{kerbl20233d} typically exhibit stronger robustness and generalization due to simplified color modeling. 
By capturing spherical variations in color and lighting, these methods reduce the reliance on scene-specific properties such as surface normals or lighting distributions. 
This simplicity allows them to operate effectively across diverse scenes without extensive parameter tuning or staged training. 
Moreover, the SH representation naturally aligns with the ``massive small particles and rapid blending'' strategy of Gaussian Splatting~\citep{kerbl20233d}, enabling efficient local approximations with low computational overhead~\citep{sloan2002precomputed}.
To address the challenge of rendering reflective scenes, we propose enhancements within the SH-based shading framework, which offers better generalization than complex PBR methods. 
Our objective is to balance generalization with the physically detailed color modeling compared with PBR-based approaches.
% In practice, we propose to decompose the rendering process into two distinct components: the transmitted component and the reflected component.
% Therefore, we seek to address the challenge of rendering reflective scenes by exploring improvements within the SH-based simple shading framework, which offers better scene generalization compared to complex PBR methods.
% Our goal is to strike a balance between generalization and the physically detailed color modeling provided by more complex PBR-based approaches.

% In this study, we introduce Ref-Unlock, a novel disentangled method that enhances the fidelity of 3DGS in accurately representing physical reflections.
In this study, we propose Ref-Unlock, a geometry-aware disentanglement framework designed to improve the fidelity of 3DGS by explicitly modeling and separating physical reflections from true scene geometry.
% Specifically, we propose to decompose the scene representation into two distinct branches: a transmitted branch and a reflected branch, each associated with its own color and opacity parameters.
Specifically, our approach separates the scene representation into two components: one modeling transmitted light and the other capturing reflected light, with each branch having dedicated color and opacity attributes.
To explicitly quantify reflection intensity, we introduce a reflection map.
% and a reflection map is introduced to explicitly indicate the reflection proportion.
To better capture high-frequency view-dependent effects, we employ the high-degree spherical harmonics for both reflected and transmitted color components, enabling finer modeling of reflective details.
% However, purely decomposing the scene without external supervision can lead to incomplete separation.
However, unsupervised scene decomposition may lead to incomplete separation.
To mitigate this, we integrate a Reflection Removal Module (RRM) based on DSRNet~\citep{hu2023single}, generating pseudo reflection-free images that guide the learning of the transmitted component and promote the disentanglement between reflection and transmission.
In addition, since reflections are inherently view-dependent and closely tied to geometry, we employ Depth Anything v2~\citep{yang2024depth} to provide pseudo-depth maps as additional supervision for the depth parameter.
To further regularize the geometry and the decomposition process, we introduce a geometry-aware bilateral smoothness constraint~\citep{guo2022nerfren}, which jointly considers both depth variations and color differences.
This ensures coherent depth prediction and more accurate reflection-transmission separation.
% This ensures that both spatial and appearance discontinuities are considered, leading to more coherent depth prediction and more accurate reflection-transmission separation.
Through the combined design of reflection-specific modeling, reflection-guided disentanglement, and geometry-aware regularization, Ref-Unlock achieves efficient and faithful reconstruction of reflective scenes without the need for additional Gaussians.
Moreover, the explicit reflection representation enables flexible editing operations such as reflection manipulation shown in Fig.~\ref{fig:teaser}.

Our contributions can be concluded as follows:
\begin{itemize}
\item We propose a dual-branch representation that separates reflected and transmitted components, together with higher-degree spherical harmonics for both branches, enabling more accurate modeling of high-frequency reflections.
\item We integrate a reflection removal module to provide pseudo reflection-free supervision, facilitating the disentanglement between reflection and transmission without requiring manual masks.
\item We leverage pseudo-depth supervision from Depth Anything v2 and a bilateral smoothness constraint that jointly regularizes depth and color, improving geometric consistency and enhancing the stability of scene decomposition.
\item Our method achieves faithful rendering of reflective scenes while maintaining high generalization to diverse environments, and enables additional scene reflection editing capabilities through the explicitly modeled reflection.
\end{itemize}

\section{Related Work}
\label{sec:Related}

\subsection{Neural Scene Representation for Novel View Synthesis}
Novel View Synthesis (NVS) aims to generate realistic images from unseen viewpoints by utilizing multiple calibrated photos that encode the scene’s 3D layout and visual details~\citep{wang2023sparsenerf,chan2023generative}.
NeRF~\citep{mildenhall2021nerf} represents a significant advance by employing neural networks to predict color and density values at sampled 3D locations along camera rays. These predictions are then combined through volume rendering to produce novel views.
However, it requires extensive sampling and numerous neural network evaluations, which make the process computationally heavy and slow, limiting its practicality for real-time applications~\citep{xu20223d}.
This challenge has led to the exploration of point-based rendering methods, which offer a better balance between rendering quality and computational efficiency~\citep{kopanas2021point,ruckert2022adop}. 
One noteworthy method that has attracted considerable interest is 3DGS~\citep{kerbl20233d,fei20243d}, which models scenes using anisotropic Gaussian functions in 3D space. 
Each Gaussian element encodes opacity and utilizes Spherical Harmonic coefficients to capture view-dependent color variations.
Unlike traditional ray tracing approaches, 3DGS leverages a point-based rasterization pipeline accelerated by modern hardware, enabling both rapid and visually detailed rendering.

While NeRF and 3DGS represent two distinct paradigms of neural scene representation—implicit and explicit, respectively—they both share the goal of modeling scenes with high fidelity and view-consistency.
However, both of them face limitations when applied to complex scenes involving diverse material properties, intricate geometry, occlusions, or other challenging visual phenomena~\citep{gao2022nerf,zhu2024scene,fei20243d}.
This has motivated a growing number of studies focused on extending these methods to improve rendering quality and physical realism in challenging scenarios.
For instance, Mip-NeRF~\citep{barron2021mip} improves NeRF by representing scenes at multiple continuous scales, reducing aliasing and enhancing detail while achieving faster rendering.
NeRFusion~\citep{zhang2022nerfusion} integrates NeRF with TSDF-based fusion to enable efficient large-scale reconstruction and photo-realistic rendering, addressing the scalability limitations of standard NeRF in complex indoor scenes.
SuGaR~\citep{guedon2024sugar} imposes regularizations to keep Gaussian points aligned with the surface, while GSDF~\citep{yu2024gsdf} and GaussianRooms~\citep{xiang2024gaussianroom} integrate 3DGS with implicit SDF fields for mutual supervision.
Among these efforts, one particularly important direction involves addressing novel view synthesis in scenes with reflective surfaces, where view-dependent appearance varies dramatically due to mirror-like effects, screen contents, or transparent materials.

\subsection{NeRF-Based Approaches for Reflective Scene Rendering}

% Reflections are a prevalent phenomenon in the real world, but accurately modeling them can be difficult without employing techniques such as ray tracing to simulate light paths~\citep{glassner1989introduction}.
% In Computer Graphics, the Screen Space Reflection~\citep{beug2020screen} technique has been developed to simulate reflection effects at a minimal computational expense. 
% Although these methods work well for image-based rendering, synthesizing novel views that preserve complex view-dependent effects, such as reflections, remains a significant challenge. 
% This issue is particularly significant in view synthesis tasks, where limited camera views often lead to incomplete observation of reflective surfaces.
% Yet, this important issue has received limited exploration~\citep{xu2021scalable,sinha2012image}.
%RefNeRF~\citep{verbin2022ref} effectively separates light into two components: diffuse and specular. It achieves this by utilizing a reflection-dependent radiance field to represent reflections considering the viewing direction accurately.
Reflection effects present significant challenges in neural scene representations due to their strong view-dependent properties and complex light interactions. 
Recent studies have extended NeRF to better capture reflection phenomena, aiming to enhance rendering fidelity in scenes containing reflective surfaces.
As a pioneering work, Ref-NeRF~\citep{verbin2022ref} effectively decomposes light into diffuse and specular components by leveraging a reflection-dependent radiance field, enabling accurate modeling of reflections based on the viewing direction.
NeRFReN~\citep{guo2022nerfren} divides the scene into several subregions, allowing for distinct treatment of various reflection types such as diffuse and specular reflections. 
This subdivision approach enhances the model’s efficiency in rendering intricate reflections and improves its capability to represent a wide range of reflective effects.
SpecNeRF~\citep{ma2023specnerf} proposes a learnable Gaussian-based directional encoding to better capture view-dependent variations, especially in regions near light sources.
% Following that, SpecNeRF~\citep{ma2023specnerf} introduces a Gaussian directional encoding that is adaptable through learning, allowing for a more accurate representation of view-dependent effects, particularly close to the light source.
% Taking a step further, Guo et al.~\citep{guo2022nerfren} introduced NeRFReN, an extension of NeRF designed to represent scenes with reflections accurately. 
% NeRFReN involves dividing a scene into two components: the transmitted and reflected parts. Each component is then modeled using separate neural radiance fields. 
% Given the inherent challenges of decomposing a scene in this manner, the authors leverage geometric priors and employ meticulously designed training strategies to attain plausible decomposition outcomes.
Following these developments, NeRRF~\citep{chen2023nerrf} incorporates the reflection equation and Snell’s law into NeRF’s ray tracing, unifying refraction and reflection via the Fresnel term. 
% A virtual cone-based supersampling module mitigates normal estimation errors, enhancing rendering quality with minimal computational overhead.
%NeRF-Casting ~\citep{verbin2024nerf} replaces costly MLP queries with reflection rays cast into NeRF geometry, sampling anti-aliased features and decoding them via a small MLP. This approach ensures consistent reflections and reduces the need for complex view-dependent representations.
%\textcolor{blue}{NeRF-Casting ~\citep{verbin2024nerf} replaces costly MLP queries with reflection rays cast into NeRF geometry, sampling anti-aliased features and decoding them via a small MLP, which ensures consistent reflections and reduces the need for complex view-dependent representations.}
NeRF-Casting~\citep{verbin2024nerf} replaces MLP queries with reflection rays cast into NeRF geometry, sampling and decoding features via a small MLP to ensure consistent reflections and simplify view-dependent representations.
% Despite these advances, the persistently slow rendering speed poses a significant obstacle to the broader applicability of these NeRF-based methods. 
Despite these improvements, the consistently high rendering time remains a major barrier to the widespread adoption of NeRF-based techniques.
To address this challenge of reflection modeling in NeRF, in this paper, we propose Ref-Unlock for the modeling of reflective scenes within the 3DGS framework based on reflection decoupling.

% The approach proposed in this paper is based on reflection decoupling and innovatively applied to the modeling of reflective scenes within the 3DGS framework. This strategy maintains the fast training and real-time rendering advantages of 3DGS.

\subsection{Reflection Modeling in 3D Gaussian Splatting}
% Built on the rendering efficiency and quality of 3DGS, several recent studies have explored modeling reflections within the 3DGS representation~\citep{jiang2024gaussianshader,gao2023relightable,malarzgaussian}.
% GaussianShader~\citep{jiang2024gaussianshader} applies a simplified shading function to 3DGS to enhance neural rendering in scenes featuring reflective surfaces, while also ensuring efficient training and rendering. 
% Relightable 3D Gaussian (R3DG)~\citep{gao2023relightable} assigns normal, BRDF properties, and incident light information to each 3D Gaussian point, allowing for the modeling of per-point light reflectance. 
% VDGS~\citep{malarzgaussian} combines the strengths of NeRF and 3DGS, resulting in swift training and inference, as well as effective modeling of shadows, light reflections, and transparency to generate realistic 3D objects. 
% However, it is important to note that these methods are currently limited to single objects rather than entire scenes.
Leveraging the efficient rendering and high-quality output of 3DGS, recent work has investigated reflection modeling using the 3DGS framework~\citep{jiang2024gaussianshader,gao2023relightable,malarzgaussian,liu2024mirrorgaussian}.
Some approaches~\citep{liu2024mirrorgaussian,meng2024mirror} have been developed to particularly tackle reflections from mirror surfaces
MirrorGaussian~\citep{liu2024mirrorgaussian} leverages mirror symmetry in point-cloud representations for efficient real-time rendering of mirrored environments, whereas Mirror-3DGS~\citep{meng2024mirror} utilizes a two-step training strategy to identify mirror-related components and estimate mirror geometry and camera poses.
Although these methods perform well in mirrored scenes, they rely on specifically crafted mirror masks, which limits their generality and practicality in diverse real-world scenarios.
In contrast, some methods based on physically based rendering (PBR)~\citep{gao2023relightable,liang2024gs,ye20243d} are capable of handling more general reflective scenes across a wider range of conditions.
GShader~\citep{jiang2024gaussianshader} applies simplified BRDF shading functions on 3D Gaussians to enhance neural rendering in scenes with reflective surfaces, focusing on decomposing color attributes into components such as diffuse, specular, and residual color for modeling reflective surfaces.
R3DG~\citep{gao2023relightable} extends 3DGS with embedded normals, BRDF parameters, and incident lighting, and uses point-based ray tracing with bounding volume hierarchies (BVH) for efficient visibility handling, enabling realistic relighting.
GS-IR~\citep{liang2024gs} extends 3DGS with staged optimization to jointly recover geometry, material, and lighting under unknown illumination, incorporating normal estimation and occlusion modeling within a full PBR pipeline.
3DGS-DR~\citep{ye20243d} leverages deferred shading with per-pixel reflection gradients to optimize surface normals during 3DGS for improved specular reflection rendering.
GeoSplating~\citep{ye2024geosplatting} combines 3DGS with explicit geometry for inverse rendering, enabling accurate normal estimation and differentiable physically based shading, but it only handles single-bounce specular lighting, with inter-reflections and baked shadows causing relighting inaccuracies.
However, it is worth noting that these designs are often optimized for particular datasets or illumination setups, which restricts their adaptability to a broader range of real-world scenes.

Therefore, our Ref-Unlock aims to handle diverse reflective phenomena in real-world scenes—such as mirrors, glass, and glossy surfaces—without relying on manually labeled masks, while remaining adaptable to varying lighting conditions.

\section{Ref-Unlock}

\label{sec:Methods}

\begin{figure*}[t]
    \centering
    \includegraphics[width=\linewidth]{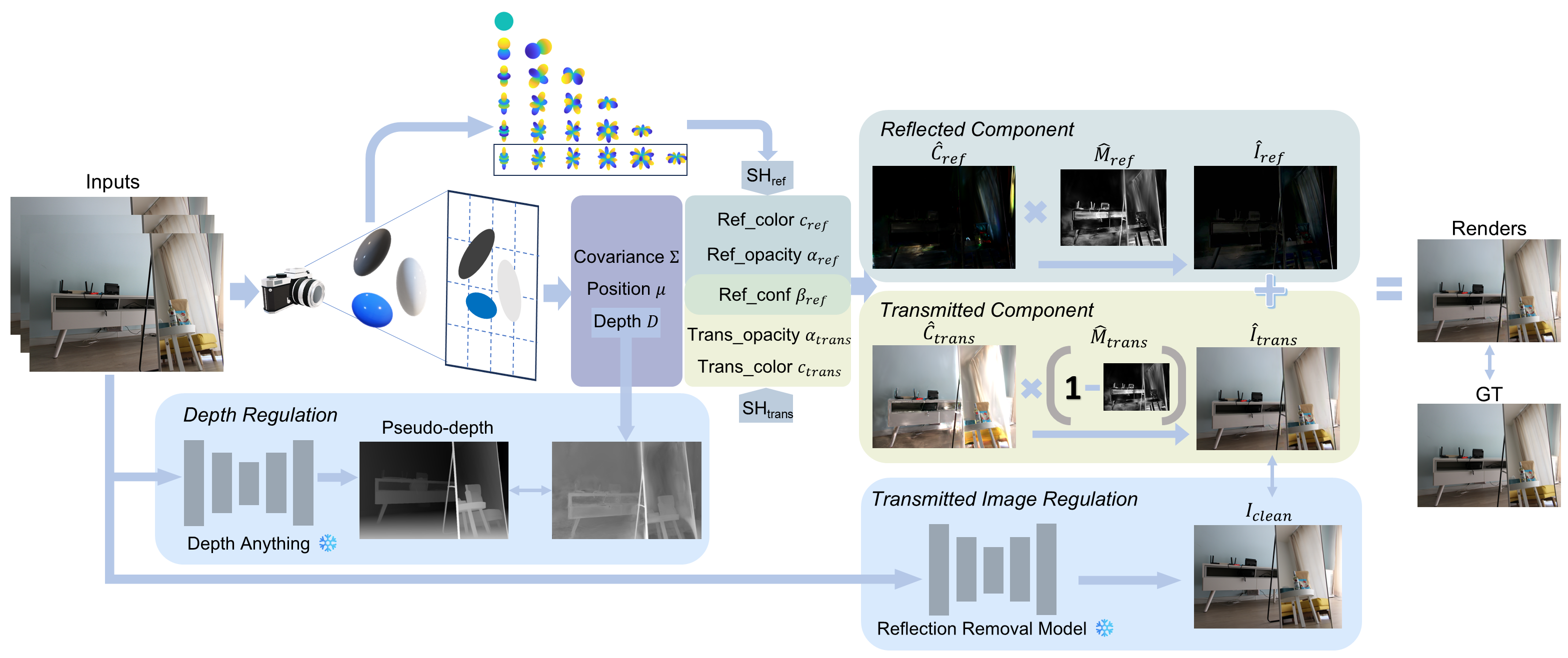}
    % \vspace{-1.3cm}
    \caption{\textbf{Overview of Ref-Unlock.} Given input images, the scene is modeled as a composition of reflected and transmitted components, each modeled with color and opacity parameters. The final rendered image is synthesized by combining the reflection and transmission branches, relying on the reflection map. }
    \label{fig:pipeline}
\end{figure*}

\subsection{Preliminary}

3D Gaussian Splatting offers an explicit formulation of radiance fields by representing the scene as a collection of anisotropic 3D Gaussian primitives~\citep{kerbl20233d,fei20243d}.
Each primitive is defined as a Gaussian function $G(X)$ over 3D space, where $X \in \mathbb{R}^3$ denotes the spatial point being evaluated:
\begin{equation}
G(X) = \exp\left(-\frac{1}{2} (X - \mu)^{T} \Sigma^{-1} (X - \mu) \right),
\label{eq:bg1}
\end{equation}
Here, $\mu \in \mathbb{R}^3$ is the mean position, and the covariance matrix $\Sigma$ characterizes the Gaussian’s spatial extent and orientation. 
To facilitate optimization, $\Sigma$ is factorized into a rotation matrix $\mathbf{R}$ and a scale matrix $\mathbf{S}$ as:
\begin{equation}
\Sigma = \mathbf{R} \mathbf{S} \mathbf{S}^T \mathbf{R}^T,
\label{eq:bg2}
\end{equation}
Practically, $\mathbf{R}$ is parameterized via a quaternion $r \in \mathbb{R}^4$, and $\mathbf{S}$ via a scale vector $s \in \mathbb{R}^3$.
Each Gaussian also carries additional attributes: an opacity value $\alpha \in \mathbb{R}$ and view-dependent appearance coefficients $\mathcal{C} \in \mathbb{R}^k$ expressed using Spherical Harmonics, where $k = (l + 1)^2$ and $l$ is the maximum SH degree.

Rendering in 3DGS is performed via a differentiable point-based splatting pipeline.
Each Gaussian is first projected onto the 2D image plane.
To efficiently render the image, the screen is divided into small tiles (e.g., $16 \times 16$ pixels), and for each tile, all Gaussians whose projections intersect with it are collected and sorted along the camera depth axis.
At each pixel, the final color $C^p$ is computed by compositing the colors and opacities of the contributing Gaussians in front-to-back order using alpha blending:
\begin{equation}
C^p = \sum_{i \in N_{\text{cov}}} c_i \alpha_i \prod_{j=1}^{i-1}(1 - \alpha_j),
\label{eq:bg3}
\end{equation}
Here, $N_{\text{cov}}$ denotes the set of Gaussians that influence the pixel, $\alpha_i$ is the effective opacity of the $i$-th Gaussian at the projected pixel location and $c_i$ is the color computed from its SH coefficients.
This compositing process ensures that closer Gaussians have a stronger influence, while more distant ones are progressively attenuated, yielding a smooth and realistic final image.
\subsection{Overview}
\par While 3DGS is effective for modeling general scenes, it struggles to model complex physical phenomena such as surface reflections. 
Specifically, 3DGS fails to explicitly model reflection components from transparent or reflective surfaces (e.g., glass or mirrors), leading to rendering inaccuracies in scenes involving such materials.
This is because reflections are inherently view-dependent and modulated by material properties, modeling their appearance using a single set of spherical harmonics is insufficient for accurate representation.
To address this limitation and improve rendering quality in reflective scenes, our Ref-Unlock extends 3DGS by introducing reflection-specific parameters that explicitly model and disentangle reflected and transmitted light components.

Firstly, as shown in Fig.~\ref{fig:pipeline}, building upon the original 3DGS parameters---mean position $\mu \in \mathbb{R}^3$, covariance $\Sigma$, and depth $\mathcal{D}\in\mathbb{R}$---our method introduces a unified parameter set that disentangles color and opacity into two branches: reflected ($c_{\text{ref}}$, $\alpha_{\text{ref}}$) and transmitted ($c_{\text{trans}}$, $\mathbf{\alpha}_{\text{trans}}$), along with a reflection map $\widehat{M}_{\text{ref}} \in [0, 1]$ that explicitly represents the proportion of reflection computed from
reflection confidence $\mathcal{\beta}_{ref}$, specifically mentioned in Sec.~\ref{sub:ref}. 
This design allows for a more comprehensive and disentangled representation of reflective scenes.

% Mentioned in Sec.\ref{subsubsec:sh} and Sec.\ref{subsub:RRM}, to better capture high-frequency reflections and make the model applicable to a wider range of scenes, we extend the spherical harmonics used for both reflected color $\mathcal{C}_{\text{ref}}$ and transmitted color $\mathcal{C}_{\text{trans}}$ from 3rd to the 5th order. 
Moreover, as discussed in Sec.~\ref{subsubsec:sh}, we increase the spherical harmonics degree from 3 to 5 for both reflected color $\mathcal{C}_{\text{ref}}$ and transmitted color $\mathcal{C}_{\text{trans}}$ to better capture high-frequency reflections and enhance the model's applicability to diverse scenes.  
This higher-degree representation facilitates more accurate modeling of complex and high-frequency view-dependent lighting effects.
To enhance the disentanglement of transmission components, we also employ an auxiliary reflection removal model~\citep{hu2023single} (Sec.~\ref{subsub:RRM}) that generates pseudo-reflection-free images $I_{\text{clean}}$, which serves as supervision for the transmitted image $\widehat{I}_{trans}$ during training.

Furthermore, considering that the view-dependent projection from 3D to 2D and rasterization process are closely related to depth parameters $\mathcal{D}$, to better model reflections that leverage view-dependent geometric properties naturally, we employ Depth Anything v2~\citep{yang2024depth} to obtain pseudo-depth maps of the scene (detailed in Sec.~\ref{sub:geo}), which supervises the depth parameters and constrain geometric features.

\subsection{Ref-Unlock with Spherical Harmonics Decomposition}
\label{sub:ref}
% However, 3DGS represents the view-dependent color information with only Spherical Harmonics coefficients, neglecting the effect of physical properties of the object in the scene, e.g., surface reflections. 
% To more comprehensively model the scenes with reflections, we extend the 3DGS with three reflection-related parameters, reflection-SH $\mathcal{C}_{ref}$, reflection opacity $\theta_{ref}$, and reflection confidence $\beta$ to construct our RefGaussian representation as follows:
\par To achieve more comprehensive modeling of reflective scenes, we decompose the original color and opacity parameters in 3DGS into two distinct branches.
Additionally, we introduce a learnable reflection confidence $\mathcal{\beta}_{ref} \in [0,1]$ to represent the reflection proportion. 
Our Ref-Unlock representations are defined as follows:
\begin{equation}\small
    \text{Ref-Unlock} = \{ \mu, \Sigma, \mathcal{D},  c_{ref}, \alpha_{ref}, c_{trans}, \alpha_{trans}, \mathcal{\beta}_{ref} \}.
    \label{eq:refgaussian}
\end{equation}
% The reflection confidence $\beta$ is a learnable parameter $\in [0, 1]$ that denotes the probability of an individual Gaussian representing the reflection part. 
% This compact and effective design allows us to decompose the rendering into two separate components, i.e., the transmitted rendering component and the reflected rendering component, with each focusing on the different aspect of the scene structure. 
% The overview of our proposed pipeline is illustrated in Figure~\ref{fig:pipeline}. 

% After projecting the Gaussians onto the 2D image plane, we conduct a synchronous rendering strategy to generate the pixel transmitted color $\widehat{C}^p_{trans}$ and reflected color $\widehat{C}^p_{ref}$ with respective parameters using Eq.~\ref{eq:bg3}. 
Once the Gaussians are projected onto the 2D image plane, we employ a unified rendering process to compute both the transmitted color $\widehat{C}^p_{trans}$ and the reflected color $\widehat{C}^p_{ref}$ for each pixel, using their respective parameters and following the compositing formulation in Eq.~\ref{eq:bg3}.

% The reflection confidence $\beta$ is accumulated in a similar way to the color rendering to produce the pixel reflection confidence $W_p$:
% Further, the pixel reflection map $\widehat{M}_{ref}^p$ is accumulated in a manner analogous to color rendering.
Further, the pixel-wise reflection mask $\widehat{M}_{ref}^p$ is computed through an accumulation process similar to that used for color composition.
Here, we use a learnable parameter reflection confidence $\beta_{ref}$ to describe it. 
The reflection confidence measures the probability that a Gaussian primitive primarily represents the reflected or transmitted part. 
This continuous weighting scheme enables differentiable separation of radiance fields while maintaining physical plausibility. 
The equation of pixel reflection map $\widehat{M}_{ref}^p$ is as follows: 
\begin{equation}
    \widehat{M}_{ref}^p = \sum_{i \in N_{\text {cov}}} \beta_{ref}^i \alpha_{trans}^i \prod_{j=1}^{i-1}\left(1-\beta^j_{ref}\right),
    \label{eq:reflection map}
\end{equation}
where $N_{\text{cov}}$ represents the set of overlapping Gaussians sorted by depth, $\beta_{ref}^i$ denotes the reflection confidence at the $i$-th Gaussian, and $\alpha_{\text{trans}}^i$ corresponds to the transmission opacity. 
The product term $\prod_{j=1}^{i-1}(1-\beta^j_{ref})$ physically represents the cumulative attenuation of reflection contribution from preceding Gaussians along the ray path. 

This formulation originates from the need to disentangle reflection and transmission effects in a physically plausible manner, where the reflection map $\widehat{M}_{\text{ref}}$ serves as the fundamental building block that generates the per-Gaussian reflection confidence parameters $\beta_{ref}^i$. The derivation follows an \textit{energy conservation principle}~\citep{hecht2002optics,pharr2016physically}. ensuring that:
\begin{equation}
\widehat{M}_{\text{ref}} + \widehat{M}_{\text{trans}} = 1.
\label{eq:6}
\end{equation}
The exponential falloff in the product term accurately models the occlusion effects between multiple reflective surfaces, while the linear combination with transmission opacity $\alpha_{\text{trans}}^i$ maintains the material-dependent relationship.

The final pixel color $\widehat{I}^p$ is obtained by fusing the transmitted and reflected components, weighted by the corresponding reflection map:
\begin{equation}
    \widehat{I}^p =  \widehat{M}_{trans}^p * \widehat{C}^p_{trans} + \widehat{M}_{ref}^p * \widehat{C}^p_{ref}.
\label{eq:7}
\end{equation}

\begin{figure}[t]
    \centering
    \includegraphics[width=\linewidth]{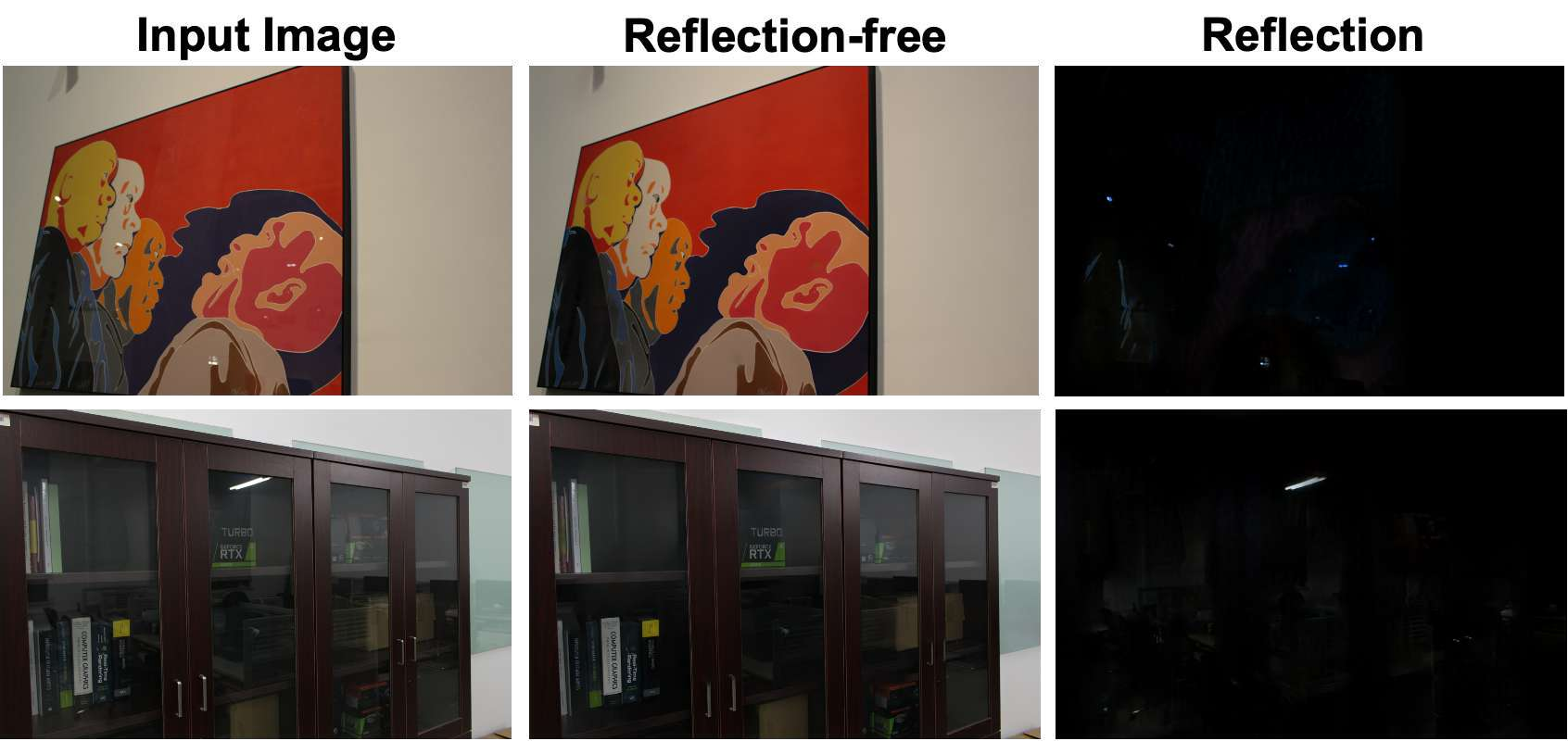}
    % \vspace{-0.7cm}
    \caption{\textbf{The results of our reflection removal module.}}
    \label{fig:reflection removal}
\end{figure}

% \begin{figure}[t]
%     \centering
%     \includegraphics[width=\linewidth]{Figures/disentanglement_v3.pdf}
%     % \vspace{-0.7cm}
%     \caption{Illustration of scene disentanglement. The final rendered images (a) combine transmitted components (b) with the product of reflected components (c) and the reflection fraction map (d).}
%     \label{fig:disentangle}
% \end{figure}

\subsubsection{High-degree Spherical Harmonics for Reflections}
\label{subsubsec:sh}
% The adoption of high-order spherical harmonics is motivated by the need to accurately represent high-frequency reflections.
The adoption of high-degree spherical harmonics is motivated by the necessity to accurately capture high-frequency lighting effects, including sharp specular reflections and fine shadowing details. 
Low-degree SH (e.g., 2 or 3 degree) is limited to encoding smooth, low-frequency signals such as diffuse lighting~\citep{green2003spherical}. 
In contrast, high-frequency reflections exhibit rapid angular variations, necessitating higher-degree basis functions for accurate reconstruction.
By employing higher-degree spherical harmonics (e.g., 5 degree or above), the basis functions achieve broader frequency bandwidth, thereby enabling accurate representation of directional light transport with high angular fidelity.
Mathematically, the approximation error of SH projections decreases exponentially with increasing degree~\citep{dai2013approximation}:
\begin{equation}
    \| L(\theta, \varphi) - \sum_{l=0}^{n}\sum_{m=-l}^{l} q_l^m Y_l^m(\theta, \varphi) \|_2 \leq A e^{-b n},
\label{eq:bg8}
\end{equation}
where $A$ and $b$ are constants dependent on the reflection sharpness, $L(\theta, \varphi)$ denotes theb target function defined over angular coordinates 
$(\theta, \varphi)$, $Y_l^m(\theta, \varphi)$ are the spherical harmonic basis 
functions of degree $l\geq 0$ and order $m$ satisfying $-l \leq m \leq l$, which can be computed using the recurrence relations of Legendre polynomials $P_l^m (x)$ as shown in Eq.~\ref{eq:bg9}, \ref{eq:bg10}, and \ref{eq:bg11}, where $K_l^m$ and $N_l^m$ are all normalization factors, and $q_l^m$ are the projection coefficients learned as parameters during the training process. The index $n$ represents the maximum degree of harmonics used in the approximation.
\begin{equation}\footnotesize
Y_l^m (\theta, \varphi) = 
\begin{cases} 
\sqrt{2} K_l^m \cos(m\varphi) P_l^m (\cos\theta) & \text{for } m > 0 \\ 
\sqrt{2} K_l^m \sin(-m\varphi) P_l^{-m}(\cos\theta) & \text{for } m < 0 \\ 
K_l^0 P_l^m (\cos\theta) & \text{for } m = 0 
\end{cases},
\label{eq:bg9}
\end{equation}
\begin{equation}\footnotesize
K_l^m = \frac{1}{N_l^m} = \sqrt{\frac{2l + 1}{4\pi} \frac{(l - |m|)!}{(l + |m|)!}},
\label{eq:bg10}
\end{equation}
\begin{equation}\footnotesize
\begin{cases}
(l - m) P_l^m (x) = x (2l - 1) P_{l-1}^m (x) - (l + m - 1) P_{l-2}^m (x)\\
P_m^m (x) = (-1)^m (2m - 1)!! (1 - x^2)^{m/2}\\
P_{m+1}^m (x) = x (2m + 1) P_m^m (x)
\end{cases}.
\label{eq:bg11}
\end{equation}

Specular reflections, particularly those resulting from sharp or glossy surfaces, exhibit rapid variations in radiance across small angular neighborhoods~\citep{ramamoorthi2001efficient}. 
Low-degree SH expansions tend to oversmooth such variations, failing to capture fine details and introducing artifacts or blurriness in the rendered image~\citep{green2003spherical}.
Using higher-degree harmonics, such as those up to the $5$ degree (resulting in 36 coefficients), can provide a better representation of high-frequency content in the reflection domain~\citep{sloan2002precomputed,green2003spherical}.

Specifically, the angular resolution of spherical harmonics increases with the max degree $n$, with each additional degree enabling the representation of finer angular variations. A $5^{\text{th}}$-degree SH expansion can effectively resolve structures with angular width on the degree of approximately $\sim \frac{180^\circ}{5+1} = 30^\circ$, making it suited to capturing the concentrated energy of glossy reflections that fall within small solid angles. Mathematically, higher-degree SH expansions are capable of approximating spherical functions with higher-frequency components, as reflected by the exponential decay in projection error shown in Eq.~\ref{eq:bg8}. In practical terms, this means that important specular features—such as tight highlights and view-dependent color shifts—can be encoded with significantly lower error, resulting in more realistic and sharper appearance in the final rendered images.

\subsubsection{Reflection Removal Module}
\label{subsub:RRM}
The transmitted component should exclusively represent the realistic (non-reflective) parts of the scene.
Most existing methods~\citep{guo2022nerfren, meng2024mirror, liu2024mirrorgaussian} rely on reflection masks to prevent interference between reflected and transmitted content.
% However, obtaining accurate reflection masks is challenging, particularly in scenes with semi-reflective overlaps. 
However, obtaining accurate reflection masks is challenging, particularly in scenes with semi-reflective overlaps, which also limits the generalization ability of existing methods.
To address this, we propose a mask-free reflection removal module to decompose the semi-reflection superpositions in the input images.
Specifically, our approach builds upon DSRNet~\citep{hu2023single}, an efficient single-image reflection removal model.
After the initialization of Gaussian primitives, the input image $I$ is processed by the reflection removal module to generate the pseudo-reflection-free image $I_{clean}$. 
Then, the rendered image $\widehat{I}$ and the transmitted part $\widehat{I}_{trans}=\widehat{M}_{trans} * \widehat{C}_{trans}$ are optimized by the photometric loss as follows:
\begin{equation}
\mathcal{L}_{\widehat{I}} = \lambda \mathcal{L}_1(I, \widehat{I}) + (1 - \lambda) \mathcal{L}_{D-SSIM}(I, \widehat{I}),
    \label{eq:12}
\end{equation}
\begin{equation}
    \begin{aligned}
        \mathcal{L}_{\widehat{I}_{trans}} &= \lambda \mathcal{L}_1(I_{clean}, \widehat{I}_{trans}) \\ &+ (1 - \lambda) \mathcal{L}_{D-SSIM}(I_{clean}, \widehat{I}_{trans}),
    % \end{align}
    \label{eq:13}
    \end{aligned}
\end{equation}
where $\lambda$ is the balance coefficient,  $\mathcal{L}_1$ calculates the absolute error between inputs while $\mathcal{L}_{D-SSIM}$ refers to the differentiable structural similarity index measure.

As shown in Fig.~\ref{fig:reflection removal}, the reflection removal module is able to effectively detect the reflection superpositions and detach them from the original images accordingly. 
By imposing the alignment with the pseudo-reflection-free image, we guide the transmitted part to focus only on the reconstruction of the realistic scene, thus achieving a better decomposition result.

Although RRM is capable of effectively separating reflection and transmission components in general scenarios, it still has several limitations. 
Specifically, the pseudo-reflection-free regions produced by RRM often exhibit blurring, leading to the loss of fine details such as textures and sharp edges in the generated clean images.
This degradation can adversely affect the geometric fidelity of the reconstructed scene.
However, in our method, RRM serves only as auxiliary supervision rather than the primary reconstruction signal. 
By integrating high-degree spherical harmonics (Sec.~\ref{subsubsec:sh}) and geometric constraints (Sec.~\ref{sub:geo}), our approach preserves high-frequency details and significantly reduces blurring artifacts.
% However, in our method, since RRM serves only as auxiliary supervision , other components such as high-order spherical harmonics mentioned in Sec.\ref{subsubsec:sh} and geometric constraints in Sec.\ref{sub: geo} can help preserve image details effectively and overcome this issue.

Moreover, in some scenes, RRM struggles to accurately determine whether objects visible in specular reflections belong to the reflected or transmitted components. 
While this issue does not substantially degrade rendering quality, it deviates from the physical principles underlying the model’s design.
We will further discuss the limitations of current one-shot approaches and explore the feasibility of future zero-shot solutions in Sec.~\ref{sec:limitation}.

\subsection{Geometry-Aware Constraints}
\label{sub:geo}

Modeling reflections in the rendering process presents significant challenges, not only because of the complexity of distinguishing their appearance from the underlying geometry but also due to the geometric ambiguities they inherently introduce.
Since the projection from 3D space to 2D image planes and the rasterization process are closely coupled with depth parameter, inaccurate depth estimation near reflective surfaces often results in distorted geometry and photometric inconsistencies.

To address these issues, we employ Depth Anything v2~\citep{yang2024depth} to generate pseudo-depth maps, which provide supervision signals for the rendered depth. 
This supervision is formulated as follows:
\begin{equation}
\mathcal{L}_{depth} = \frac{1}{H \times W} \sum_{i=1}^{H} \sum_{j=1}^{W} \left| \mathcal{D}^{pseudo}_{i,j} - \widehat{\mathcal{D}}_{i,j} \right|,
\end{equation}
where $\mathcal{D}^{pseudo}$ is the pseudo-depth predicted by Depth Anything and $\widehat{\mathcal{D}}$ is the rendered depth.

% Although effective, the pseudo-depth maps may contain local noise or edge artifacts, particularly in highly reflective regions.
% This is primarily because reflection introduces a mismatch between image appearance and scene geometry: reflective surfaces often display content from other parts of the scene rather than their own spatial structure. 
% Consequently, depth prediction models like Depth Anything, which rely on visual cues such as texture continuity and shading, may mistakenly assign incorrect depths to reflected content. 
% Additionally, abrupt color or intensity variations at the boundaries between reflective surfaces and their surroundings can be misinterpreted as depth discontinuities, introducing additional high-frequency noise in the pseudo-depth maps.

Although incorporating pseudo-depth supervision significantly enhances our model's depth perception, depth estimation models~\citep{yang2024depth} based on texture continuity and illumination cues may still assign depth values that contradict physical reality, such as assigning depth to reflected objects appearing in a mirror.
This may affect the actual placement of Gaussians and introduce noise during the optimization of their positions.
Meanwhile, as observed in the ablation study (Fig.~\ref{fig:ablation}), the depth maps generated using only pseudo-depth supervision often exhibit a lack of smooth depth transitions within certain regions (e.g., curtains in a mirror).

% To mitigate these limitations, we incorporate a depth smoothness constraint inspired by bilateral filtering, which enforces local geometric consistency guided by photometric similarity:
To mitigate the impact of physically implausible pseudo-depth and insufficient depth smoothness, we introduce a depth smoothness constraint inspired by bilateral filtering. 
This constraint enforces local geometric consistency guided by photometric similarity, helping to reduce the depth gap between reflected objects and real-world content while promoting more gradual and coherent depth transitions. 
The loss function is as follows:
\begin{equation}
    \mathcal{L}_{bi} = {\sum_{p_i{\in}\widehat{I}}}{\sum_{p_j{\in}\widehat{\mathcal{N}}_i}} f(p_i, p_j)||\widehat{D}_{p_i} - \widehat{D}_{p_j}||_1,
    \label{eq:bilateral smooth}
\end{equation}
\begin{equation}
    f(p_i, p_j) = exp(\dfrac{-||\widehat{I}^{p_i}_{trans} - \widehat{I}^{p_j}_{trans}||_1}{\gamma}),
\end{equation}
where the weight function $f(p_i, p_j)$ measures photometric similarity, and $\widehat{\mathcal{N}}_i$ denotes the 8-neighborhood of pixel $p_i$. 
Unlike NeRFReN~\citep{guo2022nerfren}, which relies on ground-truth color for weight computation, our approach directly utilizes the rendered transmitted components $\widehat{I}_{trans}$, making the constraint more tightly coupled with the learned geometry. 
This local regularization strategy effectively complements the global supervision provided by Depth Anything: while the pseudo-depth maps offer coarse yet semantically meaningful structural information, our bilateral filtering constraint simultaneously addresses local inconsistencies while preserving important geometric edges. 
The synergistic combination of these components significantly enhances both depth estimation accuracy and scene rendering consistency.

\subsection{Loss Function}

% Besides, since the rendering process is disentangled into two components, we also regularize the training of the reflected component with a reflection map smoothness:
To further stabilize the decomposition, we also add a smoothness constraint on the reflection map to reduce noise and mutual interference between the two parts:
\begin{equation}
    \mathcal{L}_{ref} = {\sum_{p_i{\in}\widehat{I}}}{\sum_{p_j{\in}\widehat{\mathcal{N}}_i}} ||\widehat{M}_{ref}^{p_i} - \widehat{M}_{ref}^{p_j}||_1,
    \label{eq:ref map smooth}
\end{equation}
where $\widehat{M}_{ref}^p$ is the rendered reflection map of the pixel. 
% With this smoothness constraint, we are able to alleviate the mutual perturbations between the two parts and present a more complete decomposition.
By introducing this smoothness constraint, interference between the two components is mitigated, leading to a more coherent and well-separated decomposition.
We optimize the attributes of Gaussians with both the photometric loss and the aforementioned constraints. 
The photometric loss $\mathcal{L}_{rgb}$ is the combination of the $\mathcal{L}_{\widehat{I}}$ and $\mathcal{L}_{\widehat{I}_{trans}}$ in Eq.~\ref{eq:12} and Eq.~\ref{eq:13}:
\begin{equation}
    \mathcal{L}_{rgb} = \lambda_{\widehat{I}} \mathcal{L}_{\widehat{I}} + S_{\widehat{I}_{trans}} * (1 - \lambda_{\widehat{I}}) \mathcal{L}_{\widehat{I}_{trans}},
\end{equation}
where $\lambda_{\widehat{I}}$ is the balance coefficient between $\widehat{I}$ and $\widehat{I}_{trans}$, while $S_{\widehat{I}_{trans}}$ represents optimization strength of transmitted branch.

% We additionally use a $\mathcal{L}_1$ loss in the first few iterations to force an alignment between the ground truth image and the transmitted image for stable training and fast convergence:
To encourage stable training and accelerate convergence, we also introduce an $\mathcal{L}_1$ loss during the initial training iterations, guiding the geometry of the transmitted component to align closely with the ground truth:
\begin{equation}
    \mathcal{L}_{init} = S_{\widehat{I}_{trans}} *\mathcal{L}_1(I, \widehat{I}_{trans}).
\end{equation}

The overall loss can be formulated as follows:
\begin{equation}
    \begin{aligned}
    \mathcal{L}_{overall} = 
    \mathcal{L}_{rgb} &+ \lambda_{init} \mathcal{L}_{init} +   \lambda_{depth} \mathcal{L}_{depth} 
    \\    &+ \lambda_{bi} \mathcal{L}_{bi} + \lambda_{ref} \mathcal{L}_{ref},
    \end{aligned}
\end{equation}
where $\lambda_{init}$, $\lambda_{depth}$, $\lambda_{bi}$ and $\lambda_{ref}$ are the coefficients of each loss term.

\section{Experiments}
\label{sec:Experiments}
\begin{table*}[t]\small
\centering
\caption{\textbf{View Synthesis Comparison Results on RFFR Dataset.} %For a comprehensive comparison, we test the performance on the whole RFFR dataset, Room from the Mip-NeRF360 dataset, and Truck from Tanks\&Temples.
}
% \resizebox{\textwidth}{!}{
\begin{tabular}{l|cccccc|c}
\toprule[1pt]
\multicolumn{1}{c}{\multirow{2}{*}{Method}}
            & art1      & \multicolumn{1}{c}{art2}        & \multicolumn{1}{c}{art3}  & \multicolumn{1}{c}{bookcase}   & \multicolumn{1}{c}{tv} & \multicolumn{1}{c}{mirror} & \multicolumn{1}{c}{\textbf{Avg.}}\\ \cline{2-8}
\multicolumn{1}{c}{} & \multicolumn{7}{c}{PSNR $\uparrow$}
 \\ \midrule
NeRF~\citep{mildenhall2021nerf}                                     &    34.686        &  40.816        & 40.304   & 29.655 & 32.863 & 32.825 &   35.191         \\

NeRFReN~\citep{guo2022nerfren}                                     &    \textbf{36.004}        &       \textbf{40.877}     &   \textbf{40.676}  &  \textbf{30.369} &  \textbf{33.306} &   \textbf{33.446} &           \textbf{35.780}     \\ \midrule
GS-IR~\citep{liang2024gs}      & 28.072 &	32.646 &	27.964 &	27.406 &	28.854 &	25.139 &	28.347   \\
GShader~\citep{jiang2024gaussianshader}         &   \underline{28.711}     &   27.198           &  30.677   & 26.930 &  30.404 &  26.521   & 28.407  \\
2D-GS~\citep{huang20242d}         &   26.664     &   33.423           &  31.537   & \underline{28.269} &  30.453 &  26.174   & 29.420   \\
3D-GS~\citep{kerbl20233d}         &   28.251     &   \underline{34.426}           &  \underline{38.732}   & 28.259 &  \underline{31.775} &  \underline{28.555}   & \underline{31.666}  \\

\textbf{Ours}  & \textbf{35.158} &   \textbf{38.405}  &    \textbf{39.747}  & \textbf{30.149} & \textbf{33.289} & \textbf{29.441}    &  \textbf{34.365} \\ \midrule
\midrule
\multicolumn{1}{c}{} & \multicolumn{7}{c}{SSIM $\uparrow$}
 \\ \midrule
NeRF~\citep{mildenhall2021nerf}    &    0.9643        &   0.9610        &  0.9591  &  \textbf{0.9233} & 0.9551 & 0.9464 & 0.9515 \\

NeRFReN~\citep{guo2022nerfren}       &      \textbf{0.9663}      &   \textbf{0.9610}   &  \textbf{0.9607}  &  0.9232  & \textbf{0.9536} & \textbf{0.9483} &     \textbf{0.9522}            \\ \midrule
GS-IR~\citep{liang2024gs}         & 0.9333 &	0.9221 &	0.8966 &	0.8854 &	0.9297 &	0.8762 &	0.9072   \\
 GShader ~\citep{jiang2024gaussianshader}        &   0.9313     &   0.8833           &  0.9180   & 0.8922 &  0.9426 &  0.9020   & 0.9116  \\
2D-GS~\citep{huang20242d}          &   0.9256     &   0.9273           &  0.9297   & 0.9099 &  0.9496 &  0.9076   & 0.9250  \\
3D-GS ~\citep{kerbl20233d}        &   \underline{0.9334}       &      \underline{0.9369}   &   \underline{0.9565}  & \underline{0.9143} & \underline{0.9501} & \underline{0.9276}     &  \underline{0.9365}   \\

\textbf{Ours}  &  \textbf{0.9684}    & \textbf{0.9535} &  \textbf{0.9593}    & \textbf{0.9192}   & \textbf{0.9575}  & \textbf{0.9351}    &  \textbf{0.9488} \\\midrule
\midrule
\multicolumn{1}{c}{} & \multicolumn{7}{c}{LPIPS $\downarrow$}
 \\ \midrule
NeRF~\citep{mildenhall2021nerf}           &    0.1859      & \textbf{0.2191}          & \textbf{0.2217}   & \textbf{0.2135} & \textbf{0.2241} & \textbf{0.1893} & \textbf{0.2089}  \\

NeRFReN~\citep{guo2022nerfren} &   \textbf{0.1828} &   0.2281  &   0.2316 &  0.2284  &  0.2306 & 0.1897 & 0.2152  \\ \midrule
GS-IR~\citep{liang2024gs}       &  0.2631 &	0.3292 &	0.3246 &	0.3015 &	0.3007 &	0.3227 &	0.3070   \\
GShader~\citep{jiang2024gaussianshader}         &   \underline{0.2361}     &   0.3896           &  0.3348   & 0.3010 &  0.2543 &  0.2889   & 0.3008  \\
2D-GS~\citep{huang20242d}          &     0.2671        &   \underline{0.3169}  &    0.3270 &  0.2886 &  0.2458 & 0.2711      & 0.2861 \\
3D-GS~\citep{kerbl20233d}         &     0.2581        &   0.3304  &    \underline{0.2681} &  \underline{0.2631} &  \underline{0.2385} & \underline{0.2373}      & \underline{0.2659} \\

\textbf{Ours}  &   \textbf{0.1621}  &      \textbf{0.2318}  &   \textbf{0.2287}  & \textbf{0.2211} & \textbf{0.2286} & \textbf{0.1944}  & \textbf{0.2111} \\ \bottomrule[1pt]
\end{tabular}
% }
\label{tab:rffr}
\end{table*}
\begin{table*}[t]\small
\centering
\caption{\textbf{View Synthesis Comparison Results on ShniyBlender Dataset.} %For a comprehensive comparison, we test the performance on the whole RFFR dataset, Room from the Mip-NeRF360 dataset, and Truck from Tanks\&Temples.
}
% \resizebox{\textwidth}{!}{
\begin{tabular}{l|ccccc|c}
\toprule[1pt]
\multicolumn{1}{c}{\multirow{2}{*}{Method}}
            & car      & \multicolumn{1}{c}{coffee}        & \multicolumn{1}{c}{helmet}  & \multicolumn{1}{c}{teapot}   & \multicolumn{1}{c}{toaster}  & \multicolumn{1}{c}{\textbf{Avg.}}\\ \cline{2-7}
\multicolumn{1}{c}{} & \multicolumn{6}{c}{PSNR $\uparrow$}
 \\ \midrule
D-NeRF~\citep{pumarola2021d}  &23.305 &26.584  &24.562 &41.628 &21.067 &28.146 \\
Ref-NeRF~\citep{verbin2022ref}  & \textbf{29.268} &	\textbf{33.617} &	\textbf{29.268} &	\textbf{44.676} &	\textbf{24.893} &	\textbf{32.345}                                            \\  \midrule
GS-IR~\citep{liang2024gs}   &  25.589 &	30.850 &	24.882 &	37.536 &	18.862 &	27.544    \\
R3DG~\citep{gao2023relightable} &  25.943 &	30.341 &	25.170 &	\textbf{43.646} &	18.776 &	28.775 \\
3D-GS~\citep{kerbl20233d}    &  \underline{25.998} &	\underline{31.224} &	\textbf{28.120} &	40.642 &	\underline{19.916} &	\underline{29.180}    \\
\textbf{Ours} & \textbf{26.823} &	\textbf{31.276} &	\underline{27.890} &	\underline{42.908} &	\textbf{21.389} &	\textbf{30.057} \\ \midrule
\midrule
\multicolumn{1}{c}{} & \multicolumn{6}{c}{SSIM $\uparrow$}
 \\ \midrule
D-NeRF~\citep{pumarola2021d}   &0.8817  &0.9312 &0.8931 &0.9923 &0.8603 &0.9117 \\
Ref-NeRF~\citep{verbin2022ref}   & \textbf{0.9514} &	\textbf{0.9711} &	\textbf{0.9514} &	\textbf{0.9951} &	\textbf{0.9052} &	\textbf{0.9548}                                           \\ 
\midrule
GS-IR~\citep{liang2024gs} &   0.8947 &	0.9532 &	0.9018 &	0.9893 &	0.7569 &	0.8992     \\
R3DG~\citep{gao2023relightable} & \underline{0.9233} &	0.9646 &	0.9314 &	\textbf{0.9954} &	0.8582 &	0.9346 \\
3D-GS~\citep{kerbl20233d}    & \textbf{0.9250} &	\textbf{0.9687} &	\textbf{0.9482} &	\underline{0.9950} &	\underline{0.8828} &	\textbf{0.9439}     \\
\textbf{Ours} & 0.9216 &	\underline{0.9654} &	\underline{0.9428} &	0.9946 &	\textbf{0.8910} &	\underline{0.9431} \\ \midrule
\midrule
\multicolumn{1}{c}{} & \multicolumn{6}{c}{LPIPS $\downarrow$}
 \\ \midrule
D-NeRF~\citep{pumarola2021d}  &0.0765 &0.1538 &0.1621 &0.1010 &0.2009 &0.1389 \\
Ref-NeRF~\citep{verbin2022ref}           & \textbf{0.0989} &	\textbf{0.0849} &	\textbf{0.0989} &	\textbf{0.0166} &	\textbf{0.1278} &	\textbf{0.0854}                                   \\ 
 \midrule
GS-IR~\citep{liang2024gs}   &  0.0814 &	0.1090 &	0.1616 &	0.0222 &	0.2388 &	0.1226    \\
R3DG~\citep{gao2023relightable} & 0.0567 &	\underline{0.0882} &	0.1238 &	\textbf{0.0120} &	0.1699 &	0.0901 \\
3D-GS~\citep{kerbl20233d} &    \underline{0.0550} &	0.0898 &	\textbf{0.0875} &	0.0138 &	0.1431 &	\underline{0.0779}      \\
\textbf{Ours} & \textbf{0.0544} &	\textbf{0.0869} &	\underline{0.1013} &	\underline{0.0136} &	\textbf{0.1272} &	\textbf{0.0767} \\ \bottomrule[1pt]
\end{tabular}
% }
\label{tab:shiny}
\end{table*}
\begin{figure*}[t]
    \centering
    \includegraphics[width=\linewidth]{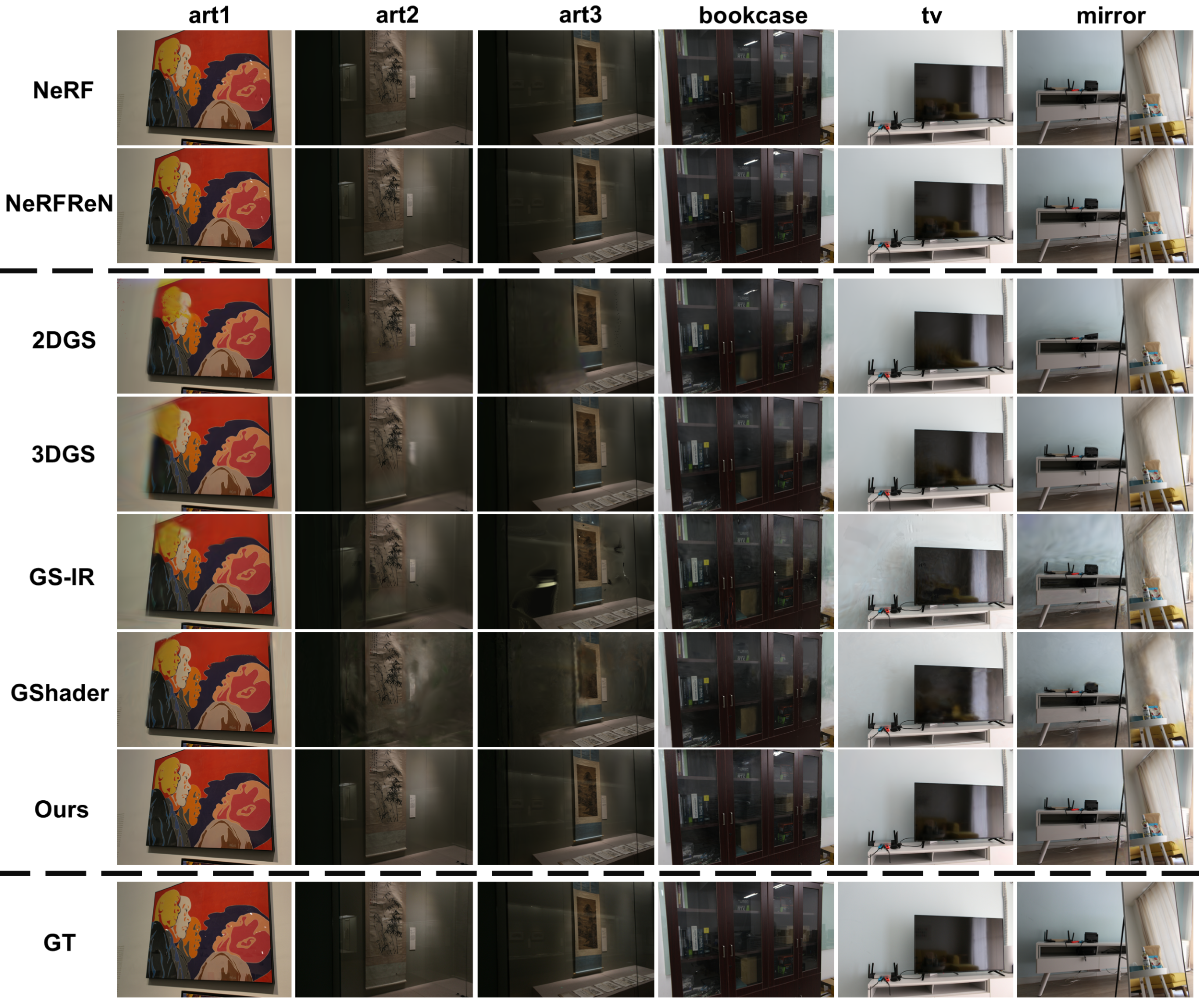}
    % \vspace{-0.7cm}
    \caption{\textbf{Visual comparisons between NeRF-based methods, GS-based methods and our Ref-Unlock.} Our method presents a more detailed and realistic rendering than GS-based methods in all cases. Compared to NeRF-based methods, our method shows comparable results in scenes with semi-reflections like \textit{art2} and \textit{art3} and delivers almost indistinguishable rendering quality in scenes containing specular reflections, like \textit{tv} and \textit{mirror}.}
    \label{fig:main}
\end{figure*}

\begin{figure*}[t]
    \centering
    \includegraphics[width=\linewidth]{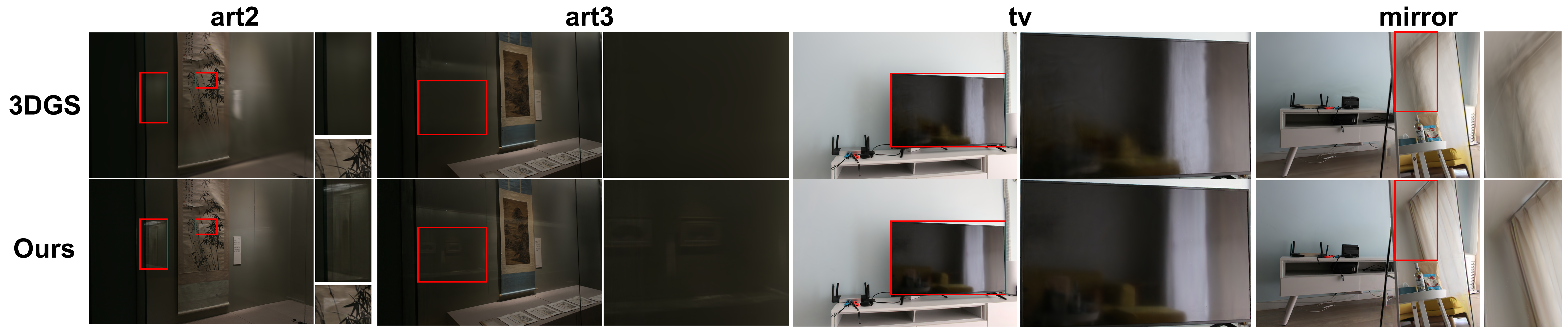}
    \vspace{-0.3cm}
    \caption{\textbf{Detailed visual comparisons between 3DGS and our method.} For a comprehensive comparison, we further show visual results of 3DGS and our method with zoom-in details on RFFR scenes.}
    \label{fig:detailed}
\end{figure*}
\begin{figure*}[t]
    \centering
    \includegraphics[width=\linewidth]{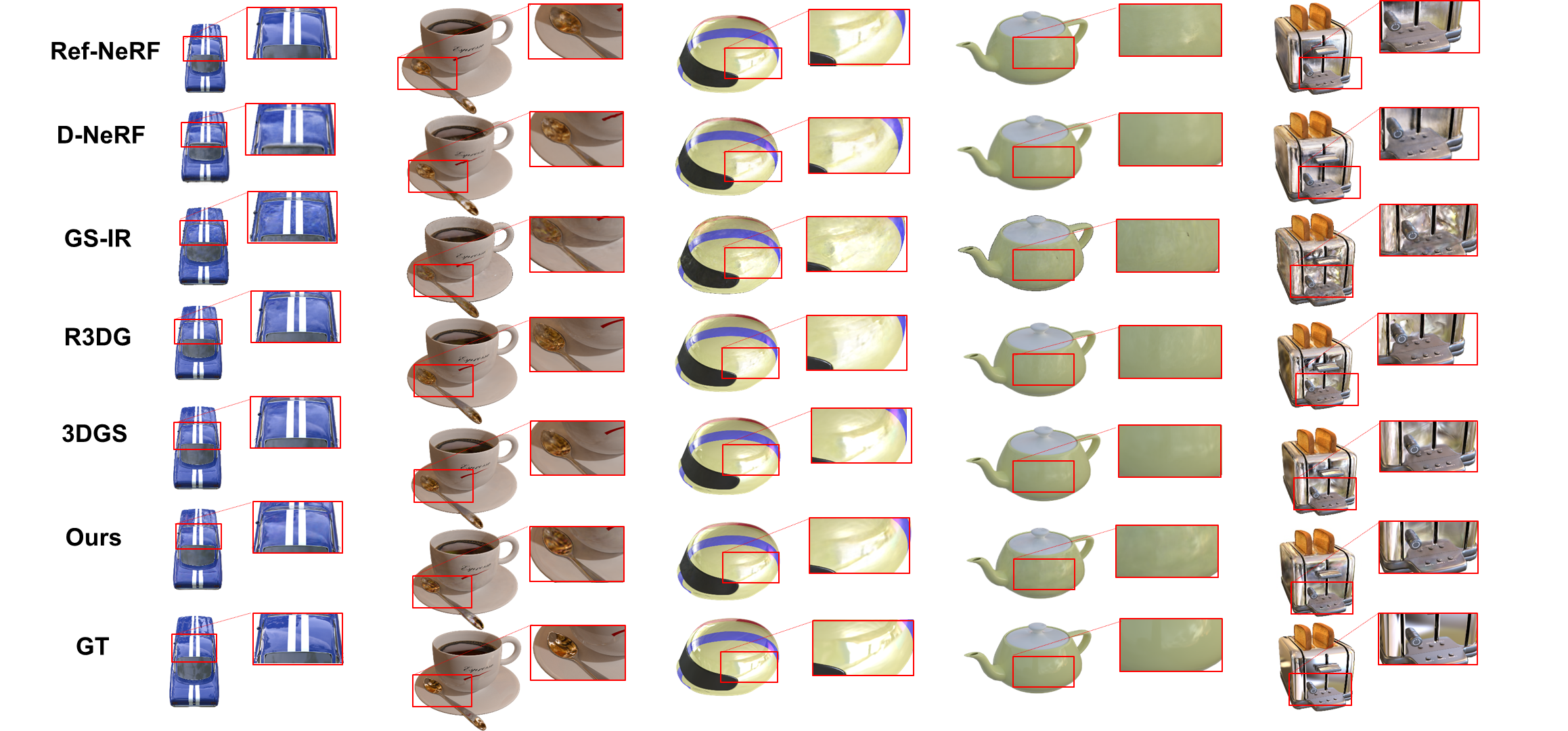}
    \vspace{-0.3cm}
    \caption{\textbf{Visual comparisons between ReFNeRF, D-NeRF, GS-IR, 3DGS, R3DG and our method,} Our method delivers visual results that not only rival those of 3DGS but, in many cases, also closely match the high visual fidelity of Ref-NeRF.}
    \label{fig:shiny_cmp}
\end{figure*}
\begin{figure*}[t]
    \centering
    \includegraphics[width=\linewidth]{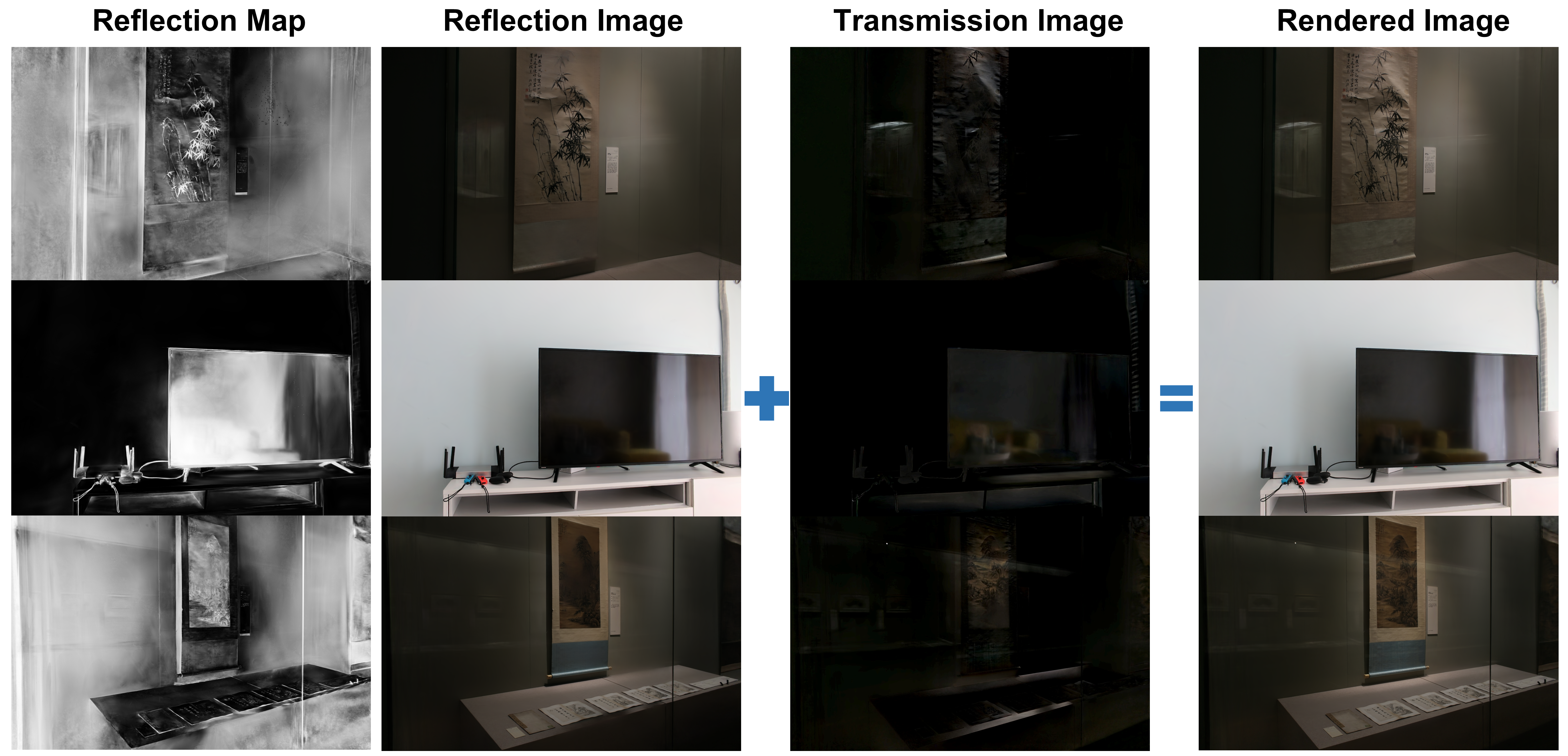}
    % \vspace{-0.3cm}
    \caption{\textbf{Reflection Disentanglement.} We present the reflection maps and the separated reflection and transmission images for three scenes: \textit{art2}, \textit{tv}, and \textit{art3}.}
    \label{fig:disentanglement}
\end{figure*}
\begin{figure*}[t]
    \centering
    \includegraphics[width=\linewidth]{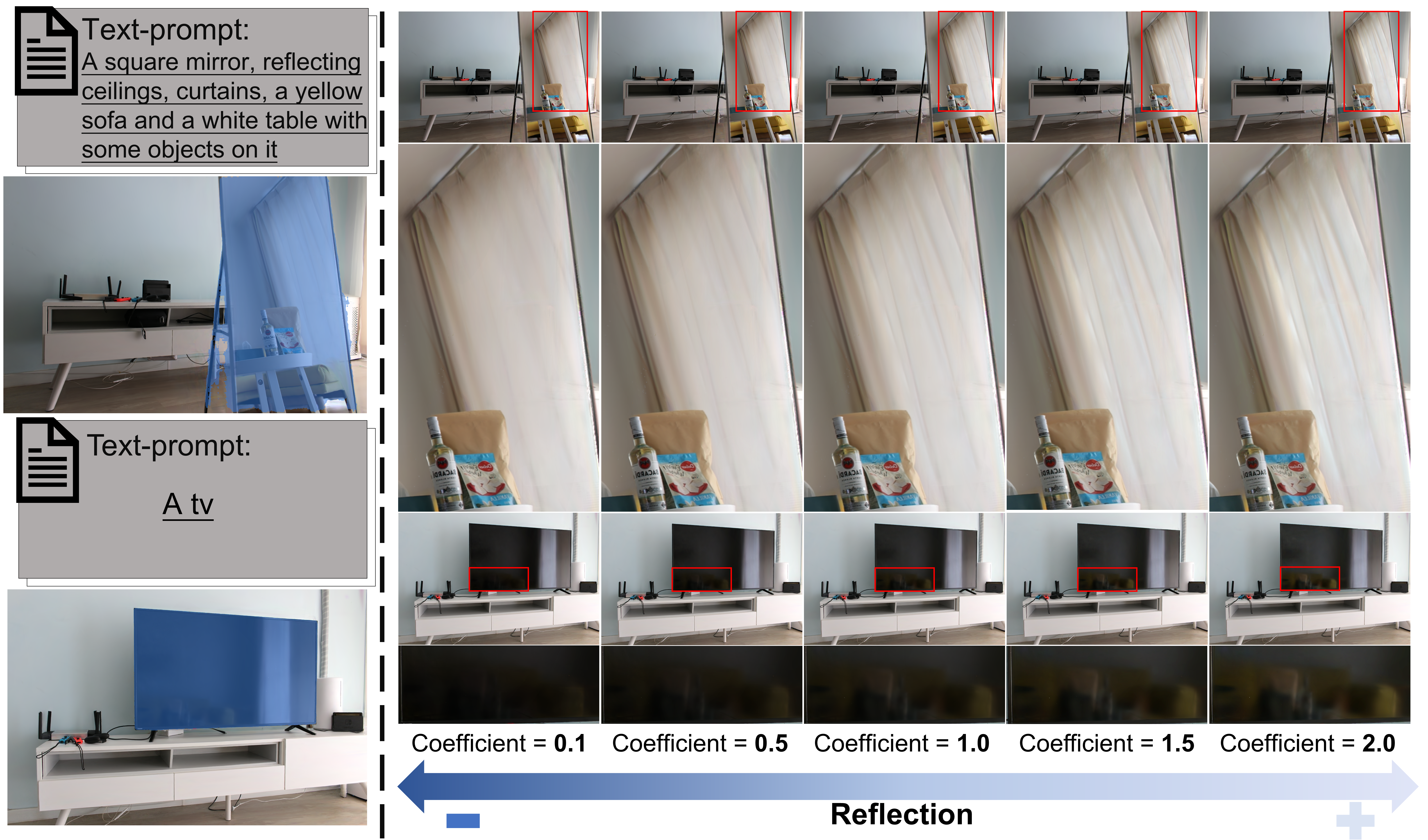}
    \vspace{-0.3cm}
    \caption{\textbf{Reflection Manipulation.} By adjusting the reflection coefficient (\textbf{0.1}, \textbf{0.5}, \textbf{1.0}, \textbf{1.5} and \textbf{2.0} from \textbf{left} to \textbf{right}) on the reflection branch, we can arbitrarily diminish or augment the brightness of the reflected content in our text-prompted regions. }
    \label{fig:manipulation}
\end{figure*}

In this section, we systematically evaluate the proposed Ref-Unlock through comprehensive experiments and analyses to demonstrate its effectiveness and versatility.

\subsection{Datasets}
%We conduct our experiments on two representative datasets to evaluate the performance of Ref-Unlock in handling reflective scenes:
We conduct experiments on two representative datasets that capture diverse reflection characteristics in both real-world and synthetic scenes:
\textbf{a)} RFFR Dataset~\citep{guo2022nerfren}: A real-world dataset specifically curated for reflection-aware rendering. 
It consists of indoor scenes with various reflective materials and surfaces, capturing complex real-world reflections under diverse lighting conditions.
\textbf{b)} Shiny Blender Dataset~\citep{verbin2022ref}: A synthetic dataset containing a range of objects with highly reflective surfaces.

\subsection{Baselines and Metrics}
We categorize the baselines into two main classes: NeRF-based and GS-based methods. 
%To ensure fairness, most comparisons are conducted within the same group, while inter-group comparisons are also provided to illustrate fundamental differences between the underlying representations.
For fairness, the majority of comparisons are performed intra-class. Additionally, inter-class comparisons are provided to elucidate the fundamental representational differences.

%The GS-based methods include both general-purpose 3D reconstruction approaches such as 3DGS~\citep{kerbl20233d} and 2DGS~\citep{huang20242d}, as well as methods specifically designed for handling reflective scenes or incorporating physically-based rendering principles, such as GShader~\citep{jiang2024gaussianshader}, R3DG~\citep{gao2023relightable}, and GS-IR~\citep{liang2024gs}. These methods leverage the efficiency and expressiveness of Gaussian primitives for scene modeling, with some incorporating additional reflection-aware or shader-level enhancements.
NeRF-based baselines include specialized methods such as NeRFReN~\citep{guo2022nerfren} and Ref-NeRF~\citep{verbin2022ref}, explicitly designed for reflective surfaces or non-Lambertian lighting scenarios. Comparative evaluation is extended to original NeRF~\citep{mildenhall2021nerf} and its dynamic extension D-NeRF~\citep{pumarola2021d}, which establish critical performance references for neural rendering benchmarks.

For GS-based methods, our evaluation includes vanilla 3DGS~\citep{kerbl20233d}, geometry-aware 2DGS~\citep{huang20242d}, and specialized methods for reflective scene modeling and physically based rendering (PBR) integration—including GShader~\citep{jiang2024gaussianshader}, R3DG~\citep{gao2023relightable}, and GS-IR~\citep{liang2024gs}. 
These methods leverage the efficiency and expressiveness of Gaussian primitives for scene representation, with some incorporating reflection-aware components or shader-level enhancements to improve rendering accuracy and visual fidelity.

%On the other hand, the NeRF-based baselines include specialized methods like NeRFReN~\citep{guo2022nerfren} and Ref-NeRF~\citep{verbin2022ref}, which are designed specifically for datasets with reflections or challenging lighting conditions. We also compare against more general NeRF-based methods such as the original NeRF~\citep{mildenhall2021nerf} and D-NeRF~\citep{pumarola2021d}, which serve as strong foundational baselines for neural rendering tasks.

%We evaluate the reconstruction quality using PSNR, SSIM, and LPIPS.
The quantitative evaluation of NVS employs three metrics: PSNR, SSIM, and LPIPS. PSNR measures pixel-level fidelity, whereas SSIM assesses structural similarity. In contrast, LPIPS evaluates perceptual similarity using deep features, offering better alignment with human visual perception. Additionally, qualitative visual comparisons are provided to facilitate an intuitive evaluation of rendering quality.

\subsection{Implementation Details}
All experiments are conducted on NVIDIA A100 GPU(s). 
For the baseline methods, we follow either the officially released settings on the corresponding datasets or use the default hyperparameters when no configuration is available for the particular dataset. 
%Our model is trained for a total of 30,000 epochs. 
Our model is optimized for a total of 30,000 iterations. 
The key hyperparameters are set as follows: $\lambda_{\widehat{I}} = 0.8$, $S_{\widehat{I}{\text{trans}}} = 10$, $\lambda_{\text{init}} = 0.01$, $\lambda_{\text{depth}} = 30$, and both $\lambda_{\text{bi}}$ and $\lambda_{\text{ref}}$ are set to 0.001. The degree of the spherical harmonics is set to 5.

\subsection{Comparisons}
%We evaluate our method on two datasets featuring reflective scenes: the RFFR real-world indoor dataset and the ShinyBlender synthetic dataset. 
%We compare Ref-Unlock against several representative GS-based and NeRF-based baselines. 
%Both quantitative results and qualitative visualizations are provided below to demonstrate our performance.

Ref-Unlock is evaluated on the real-world RFFR Dataset~\citep{guo2022nerfren} and synthetic Shiny Blender Dataset~\citep{verbin2022ref}. Extensive quantitative and visual comparisons against state-of-the-art GS-based and NeRF-based baselines validate the effectiveness and performance advantages of the proposed method.

\subsubsection{Comparisons on RFFR Dataset}
\label{sec:comparisons}
\textbf{Quantitative Results.} 
As shown in the Table~\ref{tab:rffr}, Ref-Unlock achieves the best performance among all GS-based methods in terms of PSNR, SSIM, and LPIPS. 
Compared to NeRF-based methods, although the average PSNR and SSIM of our method do not surpass the top-performing NeRF-based models, Ref-Unlock outperforms NeRFReN—the method specifically proposed for this dataset—in terms of LPIPS, indicating better perceptual quality.
Moreover, on scenes such as \textit{art1}, \textit{bookcase}, and \textit{tv}, our method achieves higher PSNR than NeRF, and on \textit{art1} and \textit{tv}, it also outperforms all NeRF-based methods in SSIM. These results highlight the effectiveness of our approach in handling reflective scenes. 

Ref-Unlock performs favorably against those GS-based methods on RFFR.
While NeRF-based methods still retain a slight numerical advantage in some metrics, our Ref-Unlock offers comparable visual quality with significantly higher efficiency. This trade-off makes Ref-Unlock a compelling choice.

\noindent\textbf{Qualitative Analysis.}
We provide qualitative comparisons on all methods in Fig.~\ref{fig:main}.
Across all tested scenes, our method achieves superior performance compared to all GS-based approaches, demonstrating its robustness and effectiveness in handling reflective scenarios.
% Additionally, our method exhibits comparable performance with NeRF-based methods in scenes with semi-reflections (e.g., \textit{art2} and \textit{art3}) and scenes with specular reflections (e.g., \textit{mirror} and \textit{tv}).
Moreover, our approach achieves performance on par with NeRF-based techniques in handling semi-reflective scenes (e.g., \textit{art2} and \textit{art3}) as well as scenes featuring specular reflections (e.g., \textit{mirror} and \textit{tv}).
The results highlight the strong reconstruction ability and photorealism of our method, especially in faithfully reproducing reflective surfaces and subtle lighting effects.

In addition, we provide a detailed visual comparison with 3DGS, the best-performing GS-based baseline in terms of quantitative metrics, to further demonstrate the advantages of our method in handling reflective regions.
As shown in Fig.~\ref{fig:detailed}, in semi-reflective scenes, our method is capable of accurately reconstructing reflective highlights on the closet surface (\textit{art2}) as well as objects such as picture frames within the reflected regions (\textit{art3}).
Moreover, in specular reflection scenes, our method provides a much better reconstruction of reflected content, such as the sofa reflected in \textit{tv} and the curtains reflected in \textit{mirror}. 
In contrast, 3DGS suffers from blurred edges and artifacts in these regions.

% The result is remarkable since our method achieves both effectiveness and efficiency without mask supervision.
% In Figure~\ref{fig:detailed}, we show a detailed comparison between RefGaussian and 3DGS with zoom-in details. Evidently, our method demonstrates outstanding rendering quality in reflected areas. Moreover, in more general cases like \textit{truck} and \textit{room}, which contain only a small fraction of reflections, our method is still able to properly reconstruct the easily neglected reflections. 

\subsubsection{Comparisons on ShinyBlender Dataset }
\textbf{Quantitative Results.} 
Table~\ref{tab:shiny} shows that our Ref-Unlock achieves the best or second-best performance in terms of PSNR and LPIPS among GS-based methods and D-NeRF. Notably, Ref-Unlock outperforms all methods except Ref-NeRF on scenes with macroscopic reflections, i.e., \textit{toaster}, \textit{coffee}, \textit{car}, where reflective effects are prominent. 
For scenes like \textit{teapot}, where reflections are subtler, our method performs on par with or even surpasses other 3DGS and D-NeRF approaches.
In terms of SSIM, Ref-Unlock achieves the highest perceptual quality on the toaster scene among GS-based methods. While it lags slightly behind Ref-NeRF and 3DGS variants on a few metrics, Ref-Unlock still ranks second in average performance overall. 
More importantly, our method offers significantly faster rendering speed, making it a more efficient and practical choice for real-time or large-scale applications.

\noindent\textbf{Qualitative Analysis.}
Furthermore, we present a comprehensive visual comparison of all six evaluated methods in Fig.~\ref{fig:shiny_cmp}.
We observe that for the teapot—a scene with relatively weak reflectivity—only our method and Ref-NeRF accurately reconstruct the subtle white highlights within the enlarged sub-image.
For the remaining four highly specular objects, our method achieves visualization results that are comparable to or even surpass those of 3DGS and, in many instances, outperform Ref-NeRF in quality. Notably, it achieves this high-fidelity rendering efficiently—without requiring object masks—while maintaining competitive training times.
In each column of Fig.~\ref{fig:shiny_cmp}, the enlarged sub-images highlight regions for detailed comparison of reflections across different methods. 
Our method achieves highly accurate rendering, particularly in preserving fine details within reflective surfaces.
Specifically, for highly specular objects such as the \textit{helmet} in Fig.~\ref{fig:shiny_cmp}, our approach not only faithfully captures the reflected content but also preserves the smoothness of the object’s material after rendering, resulting in more realistic and visually compelling results compared to existing approaches.

\subsection{Reflection Disentanglement}
\label{sec:disentanglement}
% To gain deep insight into the effectiveness of our Ref-Unlock, we further conducted experiments on reflection disentanglement for view synthesis.
To thoroughly assess the reflection handling performance of Ref-Unlock, we conducted dedicated experiments on separating reflected content for novel view generation.
As indicated in Fig.~\ref{fig:disentanglement}, the reflected components are clearly separated from the transmitted components and represented separately by reflection images $\widehat{I}_{ref}$ and transmission images $\widehat{I}_{trans}$.
The final rendered images are obtained by adding the reflection images and the transmission images together.
% The final render images combine transmitted components and reflected components multiplied by the reflection fraction map using Eq.~\ref{eq.6}.
During this process, the learned reflection map captures the spatially varying reflection intensity, where higher values (white) correspond to reflective surfaces and lower values (black) denote non-reflective regions.
As shown in Eq.~\ref{eq:7}, the reflection image is obtained through the element-wise product of the reflection map $\widehat{M}_{ref}^{p}$ and reflection color $\widehat{C}^p_{ref}$, while the transmission image is computed analogously using the transmission map $\widehat{M}_{trans}^{p}$ and transmission color $\widehat{C}_{trans}^{p}$.

% Unlike NeRFReN, which depends on manually annotated reflection masks as auxiliary input, all the images in this experiment can be directly obtained by saving the corresponding variables during the inference process of Ref-Unlock, without any additional assumptions.
% The results presented in Fig.~\ref{fig:disentanglement} demonstrate that Ref-Unlock effectively achieves transmission-reflection decomposition that aligns with human perception while maintaining high-quality view synthesis.
In contrast to NeRFReN~\citep{guo2022nerfren}, which requires manually provided reflection masks as auxiliary input, our approach generates all necessary outputs—such as transmitted and reflected components—automatically during inference, without relying on extra assumptions or supervision.
As shown in Fig.~\ref{fig:disentanglement}, Ref-Unlock produces a perceptually plausible separation of transmission and reflection, while still preserving high-fidelity novel view rendering.

% The reflection fraction map acquired through learning provides a rough indication of the surface material, with larger values (white) representing reflectors and smaller values (black) representing non-reflectors. 
% It is important to note that our primary focus is on modeling the stable virtual image through the reflected field while leaving the low-frequency highlights to the view dependency within the transmitted field. 
% Accurately estimating the position of transparent surfaces, such as glass, poses challenges due to the absence of easily traceable image features. 
% In cases like up-row in Figure~\ref{fig:disentangle}, Ref-Unlock treats the opaque surface behind as the reflective surface, which is not strictly accurate, but it does not significantly impact the quality of view synthesis due to image-space composition. 
% In more complex scenarios like bottom-row in Figure~\ref{fig:disentangle}, RefGaussian also performs well without any plane or mask assumptions, demonstrating the versatility of RefGaussian.
% Although our experiments already demonstrate impressive performance in reflection disentanglement, it is important to note that our primary focus is on modeling the stable virtual image through the reflected field, while leaving the low-frequency highlights to the view dependency within the transmitted field.
While our experimental results already showcase strong performance in separating reflections, it is worth emphasizing that our method primarily aims to reconstruct stable virtual images via the reflected component, whereas the low-frequency specular highlights are handled as view-dependent effects within the transmitted branch.
This design choice, however, may deviate from the physical laws of reflection in certain cases. For example, when dealing with mirror-like reflections, some low-frequency objects reflected in the mirror may be incorrectly attributed to the transmission component rather than the reflection component.
While this approximation is not fully physically accurate, its impact on final view synthesis quality remains negligible due to our reflection-transmission composition strategy. Future work could explore more physically precise disentanglement methods.

\subsection{Reflection Manipulation}
\label{sec:reflection manipulation}
Given the inherent advantage of our method in separating reflection and transmission regions, we can easily enable editing of the reflection intensity across the scene. 
Considering that user requirements may vary—and that global edits may not always be desirable—we further introduce a text-prompt-based region selection mechanism, allowing for more flexible and targeted reflection adjustments.

Our method allows users to define the target editing region via a text prompt. Using the EVF-SAM model~\citep{zhang2024evf}, we generate a segmentation mask for the specified region. The reflection edit is then applied by scaling the reflection branch output within the masked area by a user-defined coefficient—values greater than 1 amplify reflections, while values below 1 attenuate them. Regions outside the mask remain unmodified, ensuring the rest of the scene retains its original appearance.

As a result, we achieve editable scene reflections. As shown in Fig.~\ref{fig:manipulation}, our editing strategy effectively adjusts the reflection intensity in the scene. 
For example, the curtains reflected in \textit{mirror} and the sofa reflected in \textit{tv} become progressively brighter as the reflection coefficient increases.

Importantly, the edited scenes remain free of color artifacts or visual inconsistencies, which further demonstrates the effectiveness of our model in accurately separating reflection and transmission components.

% Unlike most methods that realize scene editing with the help of the reflection masks~\cite{guo2022nerfren}, RefGaussian conducts reflection manipulation in a more simple and straightforward manner by multiplying the reflection fraction map by a relighting coefficient that controls the intensity. 
% Such manipulation is achieved by simulating the principles of light reflection and absorption in the real physical world instead of directly operating on the image pixels. As is shown in Figure~\ref{fig:manipulation}, we can adjust the brightness of the reflections at the arbitrary scale by controlling the coefficient. Both the highlighting and the diminishing process are smooth and realistic as if the the transmission of the lights changes as the coefficient scales. 
% \textbf{Reflection Reduction}

% \textbf{Reflection Increment}

\begin{table*}[t]
\centering
\caption{Ablation Studies on Model Design.}
\setlength{\tabcolsep}{5pt}
% \resizebox{\linewidth}{!}{
\begin{tabular}{l|cccc|ccc}
\toprule[1.5pt]
  Model & Pseudo Depth  & RRM & Depth Smoothness  & Ref-map Loss   & SSIM $\uparrow$ & PSNR $\uparrow$ & LPIPS $\downarrow$ \\ \midrule
A  &    \multicolumn{1}{c}{\checkmark}  &     & \multicolumn{1}{c}{\checkmark}   &   \multicolumn{1}{c|}{\checkmark}    &   0.9475
    &      \textbf{34.432} & \underline{0.2113} \\
B  &     &  \multicolumn{1}{c}{\checkmark}   &  \multicolumn{1}{c}{\checkmark}    &  \multicolumn{1}{c|}{\checkmark}   &  0.9343     &  32.179     & 0.2360  \\  
C &   \multicolumn{1}{c}{\checkmark} &   \multicolumn{1}{c}{\checkmark}  &     & \multicolumn{1}{c|}{\checkmark} &   \underline{0.9488}
      &   34.049  &    0.2133\\
D &  \multicolumn{1}{c}{\checkmark}  &  \multicolumn{1}{c}{\checkmark}   &   \multicolumn{1}{c}{\checkmark}   &  & 0.9483 
     &  34.215 &   0.2124 \\  \midrule
Ours &   \multicolumn{1}{c}{\checkmark} &   \multicolumn{1}{c}{\checkmark}  &   \multicolumn{1}{c}{\checkmark} &   \multicolumn{1}{c|}{\checkmark}  &  \textbf{0.9488}     &    \underline{34.365}     & \textbf{0.2111}    \\
\bottomrule[1.5pt]
\end{tabular}
% }
% \vspace{-0.3cm}
\label{tab:albation on model design}
\end{table*}

% \begin{table*}[t]\small
% % \centering
% % \begingroup
% \caption{Ablation Studies on Model Design.}
% \label{tab:ablation-model-design}

% \resizebox{width=\textwidth}{!}{
% \begin{tabular}{l|cccc|ccc}
% \toprule[1.5pt]
% Model & Pseudo Depth & RRM & Depth Smoothness & Ref\_map Loss & SSIM $\uparrow$ & PSNR $\uparrow$ & LPIPS $\downarrow$ \\ 
% \midrule
% A & \checkmark &  & \checkmark & \checkmark & 0.9475 & \textbf{34.432} & \underline{0.2113} \\
% B &  & \checkmark & \checkmark & \checkmark & 0.9343 & 32.179 & 0.2360 \\  
% C & \checkmark & \checkmark &  & \checkmark & \underline{0.9488} & 34.049 & 0.2133 \\
% D & \checkmark & \checkmark & \checkmark &  & 0.9483 & 34.215 & 0.2124 \\  
% \midrule
% Ours & \checkmark & \checkmark & \checkmark & \checkmark & \textbf{0.9488} & \underline{34.365} & \textbf{0.2111} \\
% \bottomrule[1.5pt]
% \end{tabular}
% }
% % \endgroup
% \end{table*}

\begin{figure*}[t]
    \centering
    \includegraphics[width=\linewidth]{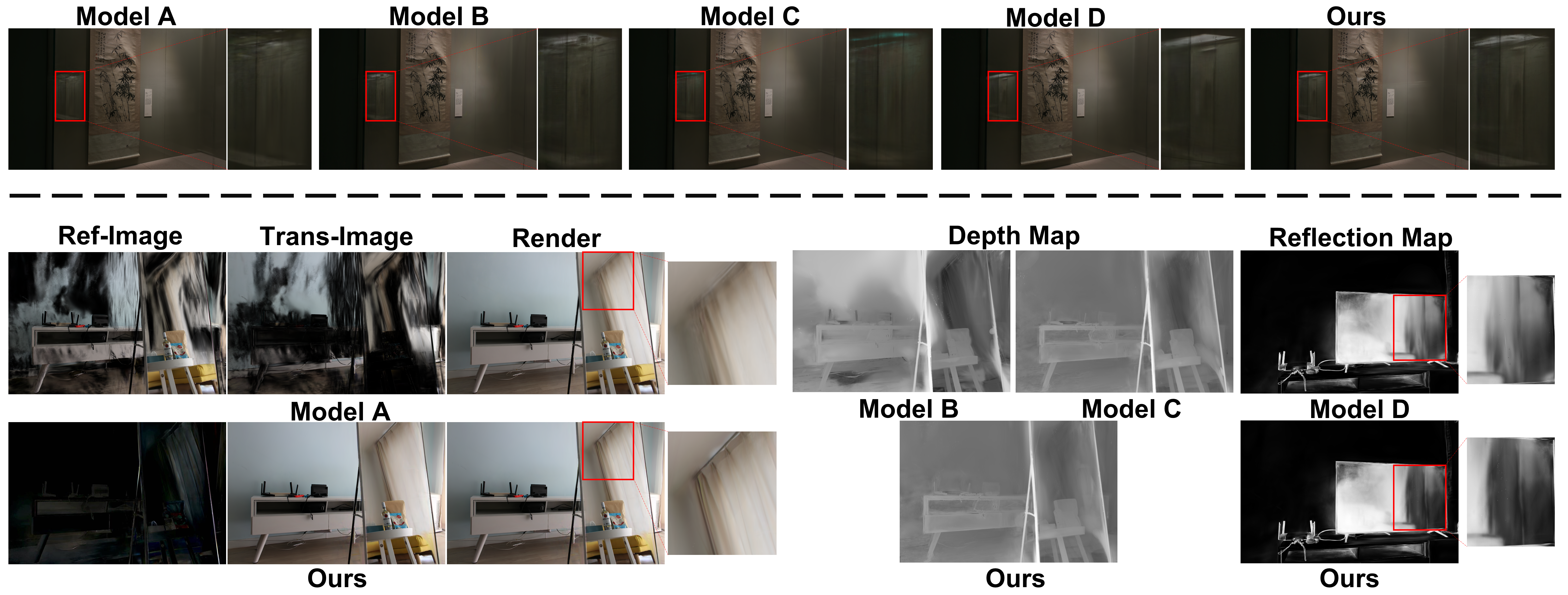}
    \vspace{-0.3cm}
    \caption{\textbf{Effectiveness of Model Design.} The combination of all model designs produces the best rendering results.}
    \label{fig:ablation}
\end{figure*}
\begin{table*}[t]\small
\centering
\caption{\textbf{Transmission Strength Ablation Results on RFFR Dataset.} %For a comprehensive comparison, we test the performance on the whole RFFR dataset, Room from the Mip-NeRF360 dataset, and Truck from Tanks\&Temples.
}
% \resizebox{\textwidth}{!}{
\begin{tabular}{l|cccccc|c}
\toprule[1pt]
\multicolumn{1}{c}{\multirow{2}{*}{\makecell{$S_{\widehat{I}_{trans}}$}}}
            & art1      & \multicolumn{1}{c}{art2}        & \multicolumn{1}{c}{art3}  & \multicolumn{1}{c}{bookcase}   & \multicolumn{1}{c}{tv} & \multicolumn{1}{c}{mirror} & \multicolumn{1}{c}{\textbf{Avg.}}\\ \cline{2-8}
\multicolumn{1}{c}{} & \multicolumn{7}{c}{PSNR $\uparrow$}
 \\ \midrule
1                                   & \textbf{35.210} &	\underline{37.720} &	 \textbf{39.895} &	 \textbf{30.525} &	 \textbf{34.035} &	 \textbf{29.957} &	 \textbf{34.557}       \\

10 (default)                                    & \underline{35.158} &	\textbf{38.405} &	\underline{39.747}	& \underline{30.149} &	\underline{33.289} &	\underline{29.441} &	\underline{34.365}      \\

20                           & 33.992 &	37.013 &	39.399 &	29.920 &	32.563 &	29.407 &	33.715         \\ \midrule
\midrule
\multicolumn{1}{c}{} & \multicolumn{7}{c}{SSIM $\uparrow$}
 \\ \midrule
1                                    &   0.9651 &	\textbf{0.9538} &	\underline{0.9593}	& \underline{0.9250} &	0.9587	& 0.9299  &     \underline{0.9486}        \\

10 (default)                                    & \textbf{0.9684} &	\underline{0.9535} &	\textbf{0.9593} &	\textbf{0.9192} &	\underline{0.9575} &	\textbf{0.9351}	& \textbf{0.9488}      \\

20                            &        \underline{0.9653} &	0.9510 &	0.9588 &	0.9120	& 0.9541 &	\underline{0.9350} &	0.9460              \\\midrule
\midrule
\multicolumn{1}{c}{} & \multicolumn{7}{c}{LPIPS $\downarrow$}
 \\ \midrule
1                                  & 0.1788 &	0.2508 &	0.2485 &	0.2289 &	\textbf{0.2282} &	0.2148 &	0.2250             \\

10 (default)                           &         \underline{0.1621} &	\textbf{0.2318} &	\underline{0.2287} &	\textbf{0.2211} &	\underline{0.2286} &	\underline{0.1944} &	\textbf{0.2111}       \\

20                                     & \textbf{0.1555} &	\underline{0.2434} &	\textbf{0.2247}	&\underline{0.2213} &	0.2323 &	\textbf{0.1909} &	\underline{0.2113}               \\ \bottomrule[1pt]
\end{tabular}
% }
\label{tab:ablation_clean}
\end{table*}
\begin{figure}[t]
    \centering
    \includegraphics[width=\linewidth]{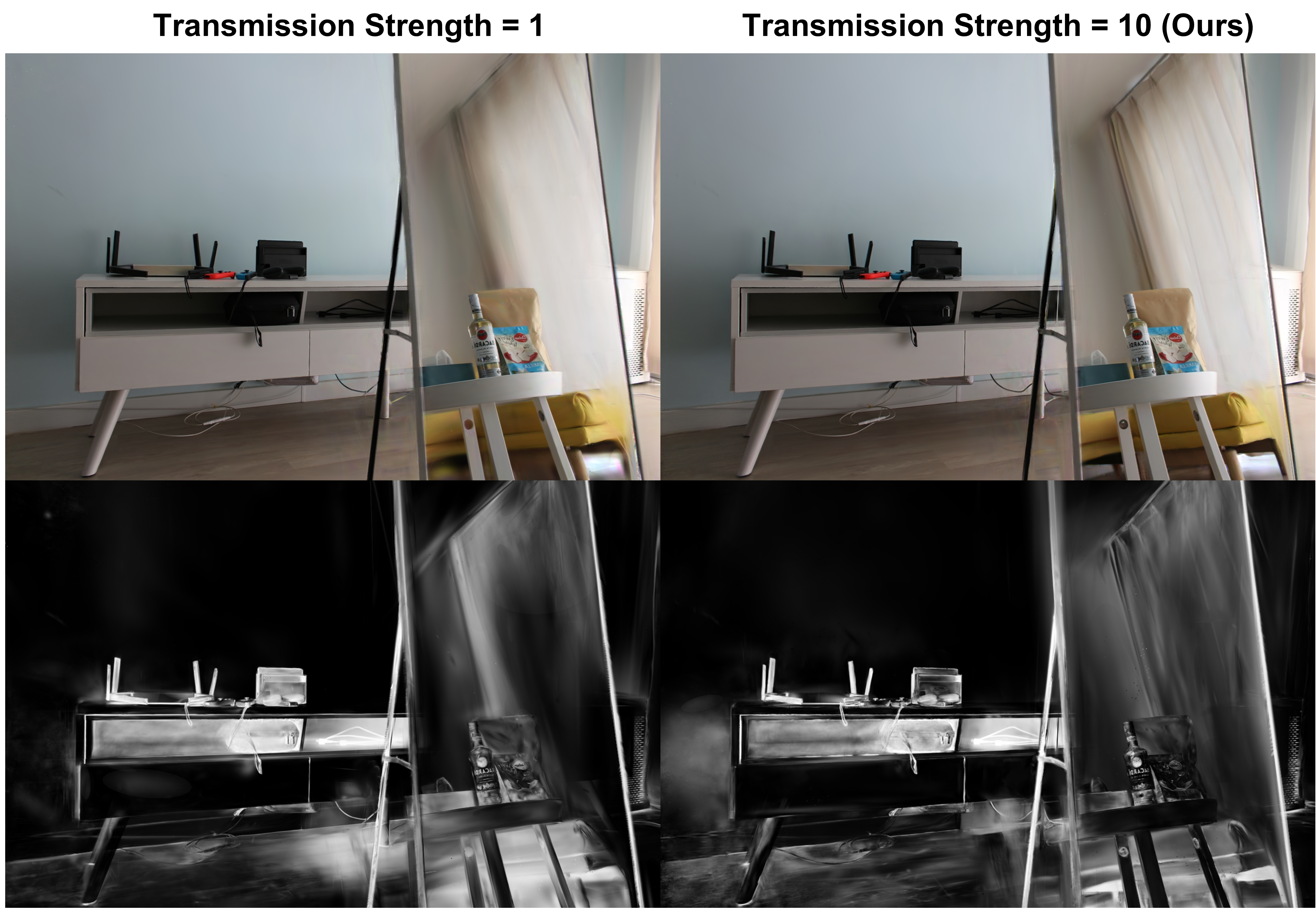}
    \vspace{-0.3cm}
    \caption{\textbf{Comparison of different transmission strengths.} We compare the visual results of models with transmission strength = 1 and 10 separately.}
    \label{fig:ablation_clean}
\end{figure}
\begin{table*}[t]
\centering
\caption{\textbf{Pseudo Depth Weights Ablation Results on RFFR Dataset.} %For a comprehensive comparison, we test the performance on the whole RFFR dataset, Room from the Mip-NeRF360 dataset, and Truck from Tanks\&Temples.
}
% \resizebox{\textwidth}{!}{
\begin{tabular}{l|cccccc|c}
\toprule[1pt]
\multicolumn{1}{c}{\multirow{2}{*}{\makecell{$\lambda_{depth}$}}}
            & art1      & \multicolumn{1}{c}{art2}        & \multicolumn{1}{c}{art3}  & \multicolumn{1}{c}{bookcase}   & \multicolumn{1}{c}{tv} & \multicolumn{1}{c}{mirror} & \multicolumn{1}{c}{\textbf{Avg.}}\\ \cline{2-8}
\multicolumn{1}{c}{} & \multicolumn{7}{c}{PSNR $\uparrow$}
 \\ \midrule
1                                  &  34.208 &	34.871 &	39.158 &	29.362 &	32.742 &	29.069 &	33.235           \\

30 (default)                                   &  \textbf{35.158} &	\underline{38.405} &	\textbf{39.747} &	\underline{30.149} &	\textbf{33.289} &	\underline{29.441} &	\textbf{34.365}       \\

50                                    & \underline{34.686} &	\textbf{38.938}	& \underline{39.614}	& \textbf{30.207} &	\underline{33.063} &	\textbf{29.507} &	\underline{34.336}               \\ \midrule
\midrule
\multicolumn{1}{c}{} & \multicolumn{7}{c}{SSIM $\uparrow$}
 \\ \midrule
1                               &    0.9657 &	0.9351 &	0.9556 &	0.9097 &	0.9547 &	0.9340 &	0.9425             \\

30 (default)                    &                 \textbf{0.9684} &	\underline{0.9535} &	\textbf{0.9593} &	\underline{0.9192} &	\textbf{0.9575} &	\underline{0.9351} &	\underline{0.9488}        \\

50                                     & \underline{0.9673} &	\textbf{0.9542} &	\underline{0.9593} &	\textbf{0.9199} &	\underline{0.9563} &	\textbf{0.9365} &	\textbf{0.9489}               \\\midrule
\midrule
\multicolumn{1}{c}{} & \multicolumn{7}{c}{LPIPS $\downarrow$}
 \\ \midrule
1                                 &   0.1706 &	0.2602 &	\underline{0.2372} &	0.2333 &	\underline{0.2293} &	0.2021 &	0.2221            \\

30 (default)                           &         \underline{0.1621} &	\textbf{0.2318} 	 & \textbf{0.2287} &	\underline{0.2211} &	\textbf{0.2286} &	\textbf{0.1944} &	\textbf{0.2111}        \\

50                                    & \textbf{0.1615} &	\underline{0.2363} &	0.2381 &	\textbf{0.2190} &	0.2295 &	\underline{0.1954} & \underline{0.2133}              \\ \bottomrule[1pt]
\end{tabular}
% }
\label{tab:ablation_depth}
\end{table*}
\begin{table*}[t]\small
\centering
\caption{\textbf{SH Degree Ablation Results on RFFR Dataset.} %For a comprehensive comparison, we test the performance on the whole RFFR dataset, Room from the Mip-NeRF360 dataset, and Truck from Tanks\&Temples.
}
% \resizebox{\textwidth}{!}{
\begin{tabular}{l|cccccc|c}
\toprule[1pt]
\multicolumn{1}{c}{\multirow{2}{*}{SH Degree $n$}}
            & art1      & \multicolumn{1}{c}{art2}        & \multicolumn{1}{c}{art3}  & \multicolumn{1}{c}{bookcase}   & \multicolumn{1}{c}{tv} & \multicolumn{1}{c}{mirror} & \multicolumn{1}{c}{\textbf{Avg.}}\\ \cline{2-8}
\multicolumn{1}{c}{} & \multicolumn{7}{c}{PSNR $\uparrow$}
 \\ \midrule
3  & \underline{34.763}	& \underline{38.701} &	39.652 &	\underline{30.116} &	33.048 &	29.220 &	34.250             \\

4  & 34.226 &	\textbf{39.121} &	\textbf{39.789} &	30.062 &	\underline{33.077} &	\underline{29.387}	& \underline{34.277}       \\

5 (default) & \textbf{35.158} &	38.405 &	\underline{39.747} &	\textbf{30.149} &	\textbf{33.289} &	\textbf{29.441} &	\textbf{34.365}               \\ \midrule
\midrule
\multicolumn{1}{c}{} & \multicolumn{7}{c}{SSIM $\uparrow$}
 \\ \midrule
3 & \underline{0.9674} &	\underline{0.9544} &	\underline{0.9593} &	\textbf{0.9214} &	\textbf{0.9579} &	\textbf{0.9368}	& \textbf{0.9495}             \\

4   & 0.9662 &	\textbf{0.9554} &	\textbf{0.9601} &	\underline{0.9201} &	0.9569 &	\underline{0.9363} &	\underline{0.9492}      \\

5 (default) &   \textbf{0.9684} &	0.9535 &	0.9593 &	0.9192 &	\underline{0.9575} &	0.9351 &	0.9488               \\\midrule
\midrule
\multicolumn{1}{c}{} & \multicolumn{7}{c}{LPIPS $\downarrow$}
 \\ \midrule
3  & 0.1657 &	0.2377 &	0.2331 &	0.2217 &	\underline{0.2309} &	\textbf{0.1940} &	0.2138            \\

4  & \underline{0.1634} &	\underline{0.2325} &	\underline{0.2300} &	\underline{0.2214} &	0.2323 &	0.1945 &	\underline{0.2124}        \\

5 (default)  & \textbf{0.1621} &	\textbf{0.2318} &	\textbf{0.2287} &	\textbf{0.2211} &	\textbf{0.2286} &	\underline{0.1944} &	\textbf{0.2111}              \\ \bottomrule[1pt]
\end{tabular}
% }
\label{tab:ablation_sh}
\end{table*}

\section{Ablation Studies}
\label{sec:ablation}

% All ablation studies are conducted on the RFFR dataset to evaluate the effectiveness of our model design and the impact of hyperparameters. Performance metrics are reported as average performance on the RFFR dataset.
We perform all ablation experiments on the RFFR dataset to assess the contribution of our model components and the influence of different hyperparameter settings. The reported results reflect the average performance across the entire RFFR dataset.

\subsection{Ablation Study on Model Design}
To assess the impact of different design choices, we trained four additional model variants to evaluate the effectiveness of the reflection removal module (RRM) constraint, pseudo depth constraint, and depth and reflection map smoothness, respectively. 
% Table~\ref{tab:albation on model design} and Fig.~\ref{fig:ablation} show the detailed setting of each model and the corresponding quantitative and visual results.
The specific configurations of each model, along with their corresponding quantitative metrics and visual comparisons, are presented in Table~\ref{tab:albation on model design} and Fig.~\ref{fig:ablation}.
The ablation study demonstrates that our fully designed model consistently outperforms the variants, achieving the highest performance.

Specifically, we observe that removing the Reflection Removal Module (RRM, Model A) results in slightly higher PSNR but lower SSIM and LPIPS compared to our full model.
More importantly, visual comparisons (left column of the second row in Fig.~\ref{fig:ablation}) suggest that the model without RRM fails to physically disentangle reflection and transmission components.
This indicates that relying solely on learning the reflection map parameters is insufficient for achieving effective separation - explicit external constraints, such as those introduced by RRM, are essential for accurate reflection disentanglement.

Moreover, removing the pseudo-depth constraint (Model B) leads to a substantial decline across all three evaluation metrics.
This validates the pivotal role of the geometry-aware constraint in enabling accurate 3D scene representation with Gaussian spheres and contributing to high-quality rendering.
As demonstrated in Fig.~\ref{fig:ablation}, the depth maps also show noticeable improvements—featuring fewer abrupt transitions and more consistent depth estimates, particularly in planar regions like walls.

The ablation study further reveals that removing either the depth smoothness constraint or the reflection map smoothness constraint (Models C and D) results in consistent performance degradation across all evaluation metrics. 
Notably, despite their relatively small weight contributions (only 0.001) in the total loss function, these constraints prove essential for optimal performance. 
In the visual comparisons (middle column of the second row in Fig.~\ref{fig:ablation}), the depth smoothness constraint leads to smoother and more natural depth transitions.
Additionally, the reflection map smoothness constraint significantly enhances boundary delineation between reflection and transmission regions through noise suppression and improved map robustness, as evidenced in the right column of the second row in Fig.~\ref{fig:ablation}.
 
\subsection{Ablation Study on Transmission Strength}

We conduct three sets of experiments with $S_{\widehat{I}_{trans}}=$ 1, 10, and 20 to decide the optimal parameter for the transmission strength.

As shown in the Table~\ref{tab:ablation_clean}, the quantitative performance of the model exhibits a rise-then-fall trend as the transmission strength increases. Notably, when the transmission strength is set to 10, the model achieves the best SSIM and LPIPS scores, and we therefore select it as the optimal parameter.

Further analysis of the other two settings reveals that, although the model with transmission strength equal to 1 attains the highest PSNR, the visual results in Fig.~\ref{fig:ablation_clean} suggest inferior disentanglement quality. Specifically, compared to the setting of 10, the reflection map under strength 1 fails to accurately distinguish between reflection and transmission regions-especially in the mirror scene, where the reflected curtain is poorly captured. This results in an under-reconstruction of fine details, manifested as noticeable blurriness in the curtain area.

For the model with a larger transmission strength (20), although the separation between reflection and transmission components improves compared to the setting of 1, the improvement over strength 10 is marginal. However, due to the increased weight, the transmission component dominates the overall loss, causing the optimization to shift its focus disproportionately toward the transmission branch. This imbalance negatively impacts the reconstruction quality of other components, such as reflection and depth, ultimately leading to a drop in overall performance.

\subsection{Ablation Study on Weight of Pseudo Depth}

Likewise, we performed three experiments to explore the impact of different weights assigned to the pseudo depth constraint.
The weight $\lambda_{depth}$ is set to 1, 30 and 50, respectively. 

As shown in the Table~\ref{tab:ablation_depth}, when the pseudo-depth loss weight is set to 30, this setting yields the highest performance with respect to both PSNR and LPIPS, and we adopt this as the final weight.
Overall, the performance also follows a rise-then-fall trend as the depth weight increases. 
This behavior essentially reflects a trade-off in model optimization.
When the weight is too low, the depth constraint is too weak, making it difficult for the model to accurately perceive and represent the scene geometry.
When the weight is too high, it dominates the optimization and impairs the learning of other components, leading to a drop in overall performance.
Therefore, a weight of 30 serves as a balanced setting that provides sufficient geometric supervision while maintaining overall model effectiveness.

\subsection{Ablation Study on Degree of SH}

Additionally, we also conduct an ablation study on the degree of SH to verify whether using higher-degree SH contributes to improved reconstruction quality of reflective scenes in our framework.

As shown in the Table~\ref{tab:ablation_sh}, increasing the SH degree consistently improves the model performance in terms of PSNR and LPIPS, indicating enhanced reconstruction accuracy and perceptual quality. 

Interestingly, we observe a slight decrease in SSIM despite improvements in PSNR and LPIPS when using higher-degree SH. 
This discrepancy can be attributed to the limitations of the SSIM metric itself~\citep{nilsson2020understanding} rather than an actual degradation in reconstruction quality. 
Specifically, SSIM is sensitive to local structural consistency but fails to capture perceptual realism in high-frequency reflective regions. 
Since reflective surfaces often contain visually accurate but structurally dissimilar content~\citep{li2025comprehensive} (e.g., mirror reflections, specular highlights), SSIM may penalize these regions even when they are faithfully reconstructed. 
In contrast, LPIPS, which aligns more closely with human perception, reflects the improved visual fidelity achieved through high-degree SH modeling.

Based on the overall evaluation results, we select the highest tested SH degree of 5 as the final configuration for our framework.
Given the substantially increased computational cost associated with higher SH degrees, and the fact that 5-degree SH already yields high-fidelity reconstruction of reflective details.

\section{Limitations and Future Works}
\label{sec:limitation}
% Since commonly seen reflectors in the real world have rather smooth structures, e.g., mirrors, windows, and screens, the proposed bilateral smoothness constraint and reflection map constraint assume the reflectors in the scene to be planar surfaces rather than irregular surfaces with significant undulations or cusps. Therefore, the applicability of our method in more diverse scenarios, e.g., scenes with curved reflectors, is currently unknown. Besides, the reflection manipulation in our RefGaussian is achieved at the pixel level through filtering the reflection map, thus producing only coarse and global enhancement. More flexible and fine-grained manipulation is a focus of our future works.  
Although our model demonstrates strong capability in separating reflection and transmission components, it currently relies heavily on the Reflection Removal Module for effective disentanglement. However, the underlying RRM (DSRNet) has inherent limitations. In certain scenarios, DSRNet fails to accurately extract the reflection layer and may introduce ambiguities by misclassifying reflected content (e.g., mirror reflections) as part of the transmission. 
Although our Ref-Unlock could correct some of these errors during optimization, such misalignment with physical reality may still misguide the training process, leading to entangled or incorrect decomposition results.

% In addition, our current illumination modeling is relatively coarse, considering only two components: reflection and transmission. It does not explicitly model the origin of reflections (e.g. inter-object reflections, point light sources or ambient lighting) or account for more accurate material-dependent interactions that light undergoes upon hitting different surfaces. This simplification may restrict the realism of the model in scenes with complex lighting conditions.

Moreover, while we employ pseudo-depth as a geometric constraint to enhance 3D scene representation, our approach remains limited to depth supervision. Key geometric attributes—such as surface normals and curvature—are not yet integrated into the learning framework. Incorporating these features could provide richer structural priors, potentially improving both the fidelity of view synthesis and the interpretability of reflection-transmission separation.

These limitations suggest several promising directions for future research: (1) developing more physically accurate and robust reflection modeling approaches, (2) enhancing geometric understanding through integration of surface normals and additional differential geometry constraints, and (3) designing illumination models that explicitly disentangle various lighting components, including inter-reflections, specular highlights, and ambient illumination.

\section{Conclusions}
\label{sec:conclusion}
% In this work, we propose Ref-Unlock, a novel 3DGS-based framework tailored for the realistic rendering of scenes with reflections.
In this paper, we introduce Ref-Unlock, a new framework grounded in 3D Gaussian Splatting that incorporates geometric awareness to enable high-fidelity rendering of reflective scenes.
% Creatively, our Ref-Unlock decomposes the scene into the transmitted component and the reflected component and fuses the renderings of two components to generate the final results.
Our Ref-Unlock framework introduces a novel strategy that separates the scene into transmitted and reflected components, and synthesizes the final output by integrating the renderings from both branches.
% Accordingly, Gaussian representation is extended with three additional parameters, namely, reflection-SH, reflection opacity, and reflection confidence to support the disentangled scheme.
Accordingly, we correspondingly increased the degree of the spherical harmonics used to represent color (radiance) in each branch, employing higher-degree spherical harmonics to better capture high-frequency reflections.
To facilitate the comprehensive decomposition of the two components, we have developed a reflection removal module to ensure the alignment of the transmitted part with scenes that are free of reflections.
Further, we constrain the model-generated depth using a pseudo-depth map to provide the model with enhanced 3D geometric awareness, thereby further improving the accuracy of view-dependent 3D scene reconstruction. At the same time, we incorporate a bilateral depth smoothness constraint to ensure that depth gradients in the scene change gradually and continuously, better reflecting the depth variation in real-world scenes.
Our method significantly outperforms the original 3DGS and some classic GS-based reflection techniques on scenes containing reflections, while delivering results comparable to NeRF-based methods. 
Finally, the ability to control and manipulate reflections highlights the potential of our approach for a wide range of practical applications.

% \newpage
% \clearpage

\backmatter

% \bmhead{Supplementary information}

% If your article has accompanying supplementary file/s please state so here. 

% Authors reporting data from electrophoretic gels and blots should supply the full unprocessed scans for key as part of their Supplementary information. This may be requested by the editorial team/s if it is missing.

% Please refer to Journal-level guidance for any specific requirements.

% \bmhead{Acknowledgements}

% Acknowledgements are not compulsory. Where included they should be brief. Grant or contribution numbers may be acknowledged.

% Please refer to Journal-level guidance for any specific requirements.

\section*{Declarations}

% Some journals require declarations to be submitted in a standardised format. Please check the Instructions for Authors of the journal to which you are submitting to see if you need to complete this section. If yes, your manuscript must contain the following sections under the heading `Declarations':

\begin{itemize}
\item Conflict of interest: Authors have no competing interests to declare that are relevant to the content of this article.
% \item Ethics approval and consent to participate
% \item Consent for publication
\item Data and code is available at our Github Repo \href{https://ref-unlock.github.io/}{https://ref-unlock.github.io/} 
% \item Materials availability
% \item Code is available at \href{https://generativediffusionprior.github.io/}{https://generativediffusionprior.github.io/}
\end{itemize}
\section*{Data Availability}
Figures~\ref{fig:main},~\ref{fig:detailed},~\ref{fig:disentanglement},~\ref{fig:manipulation},~\ref{fig:ablation},~\ref{fig:ablation_clean} and Tables~\ref{tab:rffr},~\ref{tab:albation on model design},~\ref{tab:ablation_clean},~\ref{tab:ablation_depth},~\ref{tab:ablation_sh} are experimental results on the publicly available data shared in~\cite{guo2022nerfren}. Figure~\ref{fig:shiny_cmp} and Table~\ref{tab:shiny} are experimental results on the publicly available data shared in~\cite{verbin2022ref}. All the intermediate experimental results and data shown in this paper can be accessed from the first author by request \href{https://ref-unlock.github.io/}{[https://ref-unlock.github.io/]}. 
% \noindent
% If any of the sections are not relevant to your manuscript, please include the heading and write `Not applicable' for that section. 

% %%===================================================%%
% %% For presentation purpose, we have included        %%
% %% \bigskip command. Please ignore this.             %%
% %%===================================================%%
% \bigskip

% Springer journals and proceedings: \url{https://www.springer.com/gp/editorial-policies}

% Nature Portfolio journals: \url{https://www.nature.com/nature-research/editorial-policies}

% \textit{Scientific Reports}: \url{https://www.nature.com/srep/journal-policies/editorial-policies}

% BMC journals: \url{https://www.biomedcentral.com/getpublished/editorial-policies}
% \end{flushleft}

\begin{appendices}
% \appendix

%%%%%%%%% BODY TEXT - ENTER YOUR RESPONSE BELOW

% \section{Section title of first appendix}\label{secA1}

% An appendix contains supplementary information that is not an essential part of the text itself but which may be helpful in providing a more comprehensive understanding of the research problem or it is information that is too cumbersome to be included in the body of the paper.




\end{appendices}

% \newpage
% \clearpage
% \balance

\bibliography{arxiv}% common bib file

\end{document}